\documentclass[acmlarge]{acmart}

\usepackage{multirow}
\usepackage{graphicx}
\usepackage{caption}
\usepackage{subcaption}
\usepackage{dblfloatfix}
\usepackage[table]{xcolor}
\usepackage{booktabs}
\usepackage[normalem]{ulem}
\usepackage{enumitem}
\usepackage{cleveref}
\usepackage{tabularx}
\usepackage{wrapfig}
\usepackage{algorithm}
\usepackage{algpseudocode}
\usepackage[most]{tcolorbox}
\usepackage{siunitx} 
\usepackage{tabularx}
\newcolumntype{Y}{>{\raggedright\arraybackslash}X}

\usepackage{booktabs,array}
\newcolumntype{L}[1]{>{\raggedright\arraybackslash}p{#1}}
\renewcommand\arraystretch{1.15}

\acmJournal{IMWUT}
\acmVolume{0}
\acmNumber{0}
\acmArticle{0}
\acmMonth{0}

\crefformat{section}{\S#2#1#3} 
\crefformat{subsection}{\S#2#1#3}
\crefformat{subsubsection}{\S#2#1#3}

\newcommand{\SYSTEM}{MyoText}

\begin{document}

\title{From Muscle to Text with MyoText: sEMG to Text via Finger Classification and Transformer-Based Decoding}

\author{Meghna Roy Chowdhury}
\email{mroycho@purdue.edu}
\affiliation{%
  \institution{Purdue University}
  \city{West Lafayette}
  \state{Indiana}
  \country{USA}
}

\author{Shreyas Sen}
\email{shreyas@purdue.edu}
\affiliation{%
  \institution{Purdue University}
  \city{West Lafayette}
  \state{Indiana}
  \country{USA}
}
\author{Yi Ding}
\email{yiding@pudue.edu}
\affiliation{%
  \institution{Purdue University}
  \city{West Lafayette}
  \state{Indiana}
  \country{USA}
}

\begin{abstract}

Surface electromyography (sEMG) provides a direct neural interface for decoding muscle activity and offers a promising foundation for keyboard-free text input in wearable and mixed-reality systems. Previous sEMG-to-text studies mainly focused on recognizing letters directly from sEMG signals, forming an important first step toward translating muscle activity into text. Building on this foundation, we present \SYSTEM{}, a hierarchical framework that decodes sEMG signals to text through physiologically grounded intermediate stages. \SYSTEM{} first classifies finger activations from multichannel sEMG using a CNN–BiLSTM–Attention model, applies ergonomic typing priors to infer letters, and reconstructs full sentences with a fine-tuned T5 transformer. This modular design mirrors the natural hierarchy of typing, linking muscle intent to language output and reducing the search space for decoding. Evaluated on 30 users from the \textit{emg2qwerty} dataset, \SYSTEM{} outperforms baselines by achieving 85.4\% finger-classification accuracy, 5.4\% character error rate (CER), and 6.5\% word error rate (WER). Beyond accuracy gains, this methodology establishes a principled pathway from neuromuscular signals to text, providing a blueprint for virtual and augmented-reality typing interfaces that operate entirely without physical keyboards. By integrating ergonomic structure with transformer-based linguistic reasoning, \SYSTEM{} advances the feasibility of seamless, wearable neural input for future ubiquitous computing environments.

\end{abstract}

%




\maketitle


\begin{figure}[t]

    \centering

    \includegraphics[width=0.8\linewidth]{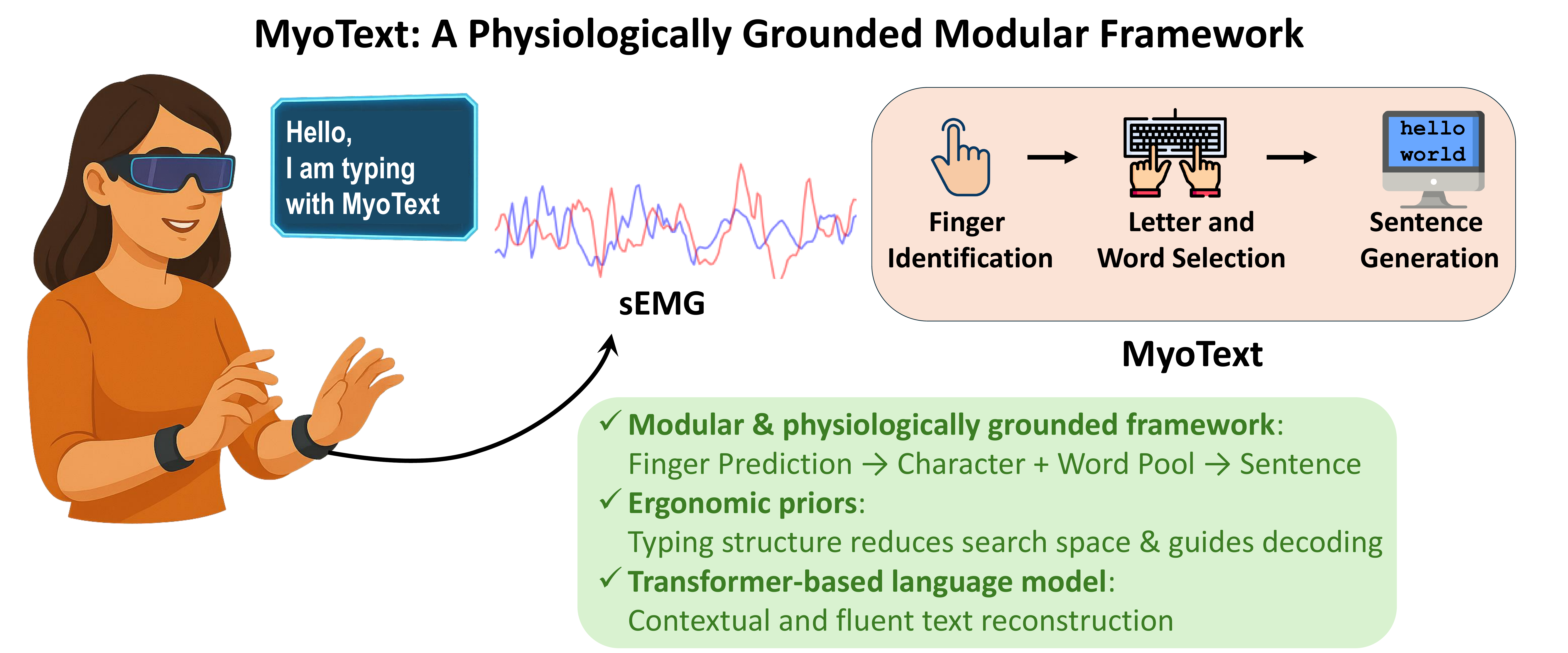}

    \vspace{-2mm}

    \caption{\textbf{\SYSTEM{}.} A physiologically-grounded modular sEMG-to-text decoding framework combining finger classification, ergonomic letter pooling, and sentence reconstruction. }

    \vspace{-3mm}

    \label{fig:intro}

\end{figure}

\section{Introduction}

Advances in wearable and ubiquitous computing are increasingly blurring the boundaries between humans and machines, enabling continuous, hands-free interaction across physical and virtual worlds~\cite{sen2024human}. As biosensing devices, which measure biological signals, grow smaller and integrate further into daily life, users need reliable, natural input methods that do not depend on specific surfaces~\cite{chatterjee2023bioelectronic}. Yet, common inputs like touchscreens, cameras, and motion sensors still rely on sight, lighting, or large gestures, limiting mobile and AR performance~\cite{rakkolainen2021technologies}.

Among new input technologies, electromyography (EMG) offers a direct link to the body’s motor system by measuring electrical signals produced when muscles move~\cite{sen2023machine}. Surface EMG (sEMG) captures these signals non-invasively through skin-mounted electrodes~\cite{chowdhury2013surface,de2002surface}, supporting quiet and flexible interaction. These properties make sEMG useful in wearable and mixed-reality devices such as smartwatches, AR glasses, and neural wristbands~\cite{sen2023machine,rani2023surface}. Recent systems like Meta Neural Band show that wrist-based sEMG sensors can recognize subtle finger motions for low-effort and discreet interactions~\cite{kaifosh2024generic,meta2021orion,OrionAIG74:online}. These capabilities are important for AR and VR text input, enabling users to type in the air without physical keyboards. In such environments, decoding muscle activity through sEMG is an essential step toward achieving seamless text entry in extended reality (XR).

Despite this promise, realizing robust sEMG-to-text systems requires addressing three key challenges. First, sEMG signals naturally vary across users and sessions due to electrode placement, muscle fatigue, and individual anatomy~\cite{shabanpour2025moemba,wang2022multi,sousa2023long,li2020analysis,li2024non}. Researchers have developed important stabilization techniques, including signal filtering, standardized electrode positioning, and per-user calibration, though achieving reliable cross-user generalization without calibration remains an open problem. Second, to handle this variability, the field has made progress through hierarchical decoding pipelines that first detect motion before classifying gestures or characters~\cite{leone2019smart,chen2023hierarchical}, and through sequence models such as RNNs~\cite{hannun2019sequence} and transformers~\cite{synnaeve2019end} that capture temporal dependencies. These architectural advances have improved within-session stability, and scaling them to large, linguistically structured datasets represents the next frontier. Third, while foundational datasets like NinaPro (27 users)~\cite{Ninapro25:online}, MyoGym (12 users)~\cite{koskimaki2017myogym}, CSL-HDEMG (5 users)~\cite{amma2015advancing}, and CapgMyo (18 users)~\cite{du2017surface}, which are established for gesture recognition, continuous text-level decoding requires larger-scale data with naturalistic language structure. The \textit{emg2qwerty} dataset~\cite{sivakumar2024emg2qwerty} from Meta Reality Labs represents a major milestone, collecting annotated sEMG signals from 108 users during typing tasks. Participants were screened to ensure at least 90\% correct finger-to-key mappings under standard QWERTY layout~\cite{TypingLe73:online,Typingwi87:online,WhereSho97:online}, capturing ergonomic typing behavior at scale. Using this dataset, the authors trained a convolutional neural network that maps sEMG activity directly to typed characters, achieving substantial character error rate (CER) improvements when augmented with a 6-gram language model. This result highlights an important insight: the strong reliance on linguistic priors suggests that direct character prediction from sEMG may not align with the physiological structure of the signal. sEMG naturally encodes finger-level muscle activations, motivating approaches that first decode fingers before resolving character ambiguity through ergonomic and linguistic constraints.

Building on this insight, we propose \SYSTEM{}, a \emph{physiologically grounded}, \emph{modular} sEMG-to-text framework that models typing as a motor–linguistic hierarchy. As shown in Fig.~\ref{fig:intro}, \SYSTEM{} proceeds through three stages: (1) finger classification from sEMG signals, (2) ergonomically constrained letter and word pooling, and (3) transformer-based sentence generation. This design is motivated by three core principles. First, sEMG signals naturally encode finger-level muscle activations, not abstract character identities. By predicting fingers rather than letters directly, \SYSTEM{} aligns the decoding task with the physiological structure of the signal, reducing ambiguity and improving cross-user generalization. Second, typing ergonomics provides structural constraints that reduce the decoding search space. The standardized QWERTY finger-to-key mapping defines a natural prior that guides letter and word pooling, while remaining robust to non-canonical typing patterns through the downstream transformer. Third, separating motor decoding from linguistic reasoning improves both accuracy and deployability. Inspired by hierarchical decoding in brain–computer interfaces~\cite{leone2019smart,chen2023hierarchical}, this modularity enables distributed execution, where the lightweight finger classifier can operate on resource-constrained wearables, while the transformer decoder can operate on resource-rich paired devices like smartphones~\cite{chowdhury2024leveraging}.

We evaluate \SYSTEM{} on the \textit{emg2qwerty} dataset with 30 participants, selected for diversity and computational feasibility. \SYSTEM{} achieves 85.4\% finger classification accuracy and reconstructs sentences with a 5.4\% Character Error Rate (CER) and 6.5\% Word Error Rate (WER), substantially outperforming prior direct character prediction approaches. We further tested robustness under non-canonical finger–key mappings and linguistic variations, observing consistent decoding results.

To summarize, this work makes the following contributions:
\begin{itemize}
    \item We introduce \SYSTEM{}, a physiologically grounded, hierarchical sEMG-to-text decoding framework that separates motor intent from linguistic decoding through finger classification, ergonomic constraints, and a transformer-based sentence generation approach.
    \item We design a CNN–BiLSTM–Attention model for finger activation recognition and a constrained T5 decoder that achieves 85.4\% finger prediction accuracy and 5.4\% character error rate on \textit{emg2qwerty}.
    \item We encode typing ergonomics as a structural prior that constrains the search space for transformer-based decoding and improves robustness to non-canonical finger–key mappings.
    \item We analyze the \textit{emg2qwerty} dataset, revealing consistent finger-dependent sEMG structure and demonstrating generalization of \SYSTEM{} through leave-one-user-out evaluation, cross-dataset testing on English corpora, and robustness under non-canonical typing patterns.
\end{itemize}
Together, these contributions advance sEMG-based text systems by showing that generalizable decoding is possible. By combining ergonomic design, modular architecture, and robust language models, this work establishes a practical and scalable foundation for sEMG-driven communication, paving the way for the future of the Internet of Bodies~\cite{rana_dac}.

\begin{wrapfigure}[14]{r}{0.4\textwidth}
  \centering
  \vspace{-0.4in}
  \includegraphics[width=\linewidth]{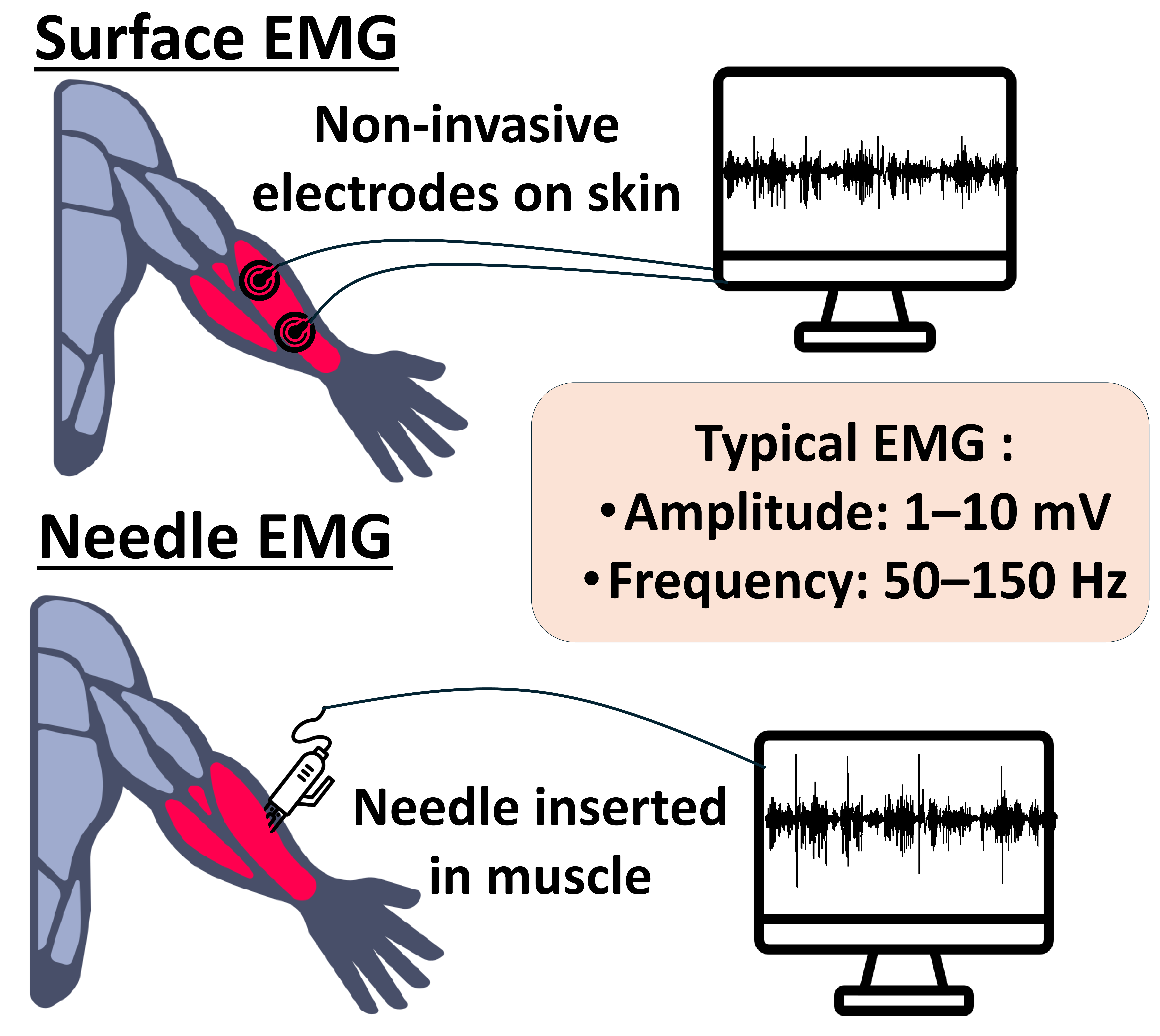}
  \caption{\textbf{EMG Sensing Modalities.} Needle and surface EMG signal acquisition methods.}
  \vspace{-0.2in}
  \label{fig:emg}
\end{wrapfigure}

\section{Background and Related Work} \label{sec:emg}

In this section, we first introduce the electromyography (EMG) biosignal and its underlying physiological principles. We then concentrate on the key applications of EMG in wearable devices, with a primary focus on sEMG-to-text decoding systems. Finally, we discuss the research gaps in current sEMGs-to-text studies.

\paragraph{\textbf{Electromyography (EMG)}}
Electromyography (EMG) is a physiological signal that captures the electrical activity produced by skeletal muscle activation, offering a direct interface to the human neuromuscular system~\cite{mills2005basics}. EMG signals are generated by motor unit action potentials and are typically in the range of 1–10 mV, with dominant frequencies between 50–150 Hz~\cite{de2002surface}. These biopotential signals can be measured either invasively using needle electrodes, or non-invasively via surface EMG (sEMG) using skin-mounted electrodes~\cite{gohel2020review}. While needle EMG offers clinical precision, sEMG is more widely adopted in wearable and mobile contexts due to its comfort, accessibility, and potential for integration with everyday interfaces~\cite{rubin2019needle}. Muscle tissue remains electrically quiescent at rest, but during contraction, it produces distinctive spatiotemporal patterns that encode motor intent~\cite{Electrom90:online}. These patterns form the basis for interpreting user actions and enabling responsive, intent-aware systems.

\paragraph{\textbf{sEMG with Wearable Devices}}

Recent advances in wearable design have significantly improved the usability of sEMG-based systems~\cite{sen2023machine}. These devices are now integrated into armbands, textiles, smart shoes, and gloves, spanning a wide range of platforms. Table~\ref{tab:emg_devices} lists representative sEMG-enabled wearables and their application domains~\cite{kartsch2018smart, maragliulo2019foot, rawat2016evaluating, feng2021active, Wearable26:online, di2020validation, finni2007measurement, leone2019smart}. Many of these devices support gesture recognition, mobility analysis, or rehabilitation, demonstrating the versatility of the sEMG signal in human-computer interaction. The latest advancement in sEMG wearables is the Meta Neural Band~\cite{OrionAIG74:online}, a neural interface that decodes fine-grained motor intent from wrist-based sEMG with low-latency responsiveness. These developments illustrate sEMG's transition from a clinical and biomechanical tool to a ubiquitous input interface, enabling next-generation typing and control paradigms.

\begin{wraptable}{r}{0.6\textwidth}
\vspace{-0.2in}
  \caption{\textbf{Summary of sEMG Wearable Devices.}}
  \small
  \label{tab:emg_devices}
  \begin{tabular}{lll}
    \toprule
    \textbf{Ref.} & \textbf{Form Factor} & \textbf{Application} \\
    \midrule
    \cite{kartsch2018smart} & Wearable Wristband & Gesture Recognition (Solar Powered) \\
    \cite{maragliulo2019foot} & Smart Footwear & Foot Gesture Recognition \\
    \cite{finni2007measurement} & Textile Shorts & sEMG Monitoring \\
    \cite{leone2019smart} & Smart Socks & Leg Muscle Assessment \\
    \cite{feng2021active} & Robotic Glove & Hand Rehabilitation \\
    \cite{Wearable26:online} & Msleeve and Mbody Suit & Sports and Ergonomics \\
    \cite{rawat2016evaluating} & Myo Armband & sEMG Sensing for Gesture Control \\
    \cite{OrionAIG74:online} & Meta Neural Wristband & Neural Interface and Input Decoding \\
    \bottomrule
  \end{tabular}\vspace{-0.1in}
\end{wraptable}

\paragraph{\textbf{Applications of sEMG}}

sEMG originated as a diagnostic tool for neuromuscular disorders, like muscular dystrophy, myasthenia gravis, and carpal tunnel syndrome~\cite{staalberg1991scanning,subasi2013classification}. It then enabled myoelectric prostheses, where residual muscle signals provide intuitive control of artificial limbs~\cite{zhou2007decoding, sudarsan2012design,cimolato2022emg}. As technology matured, sEMG expanded into sports medicine, rehabilitation, and ergonomics, using real-time muscle feedback to improve posture, reduce injury risk, and enhance performance~\cite{clarys1993electromyography, vinod2013integrated,al2023electromyography,kuo2019immediate}. Applications span wearable robotics, gait analysis, fall-risk monitoring, and studies of age-related sarcopenia~\cite{gurchiek2021wearables,freed2011wearable,nouredanesh2016detection,leone2017fall,tian2010mechanomyography}. Beyond physical applications, sEMG increasingly translates human intent into communication, where systems recognize sign-language gestures, supporting more natural interfaces for the deaf community, and assist individuals with partial paralysis or motor impairments by decoding intended movements into text or control commands~\cite{el2024design,srinivas2024empowering,booker2020future,al2024advancements,pinheiro2011alternative,choudhary2022machine}. This sEMG-to-text trajectory captures muscle activity from the forearm, face, and larynx and maps it to written language or control signals~\cite{janke2017emg,jou2006towards,wand2011session}. sEMG is also integrated into immersive virtual environments, translating muscle intent directly into digital control~\cite{toledo2022virtual,dwivedi2020emg,karen}. The latest step is Meta’s sEMG neural-interface wristband, which detects subtle muscle signals and converts them into text and commands, advancing seamless, non-invasive interfaces for everyday communication and interaction~\cite{kaifosh2024generic,meta2021orion}.

\paragraph{\textbf{Challenges with sEMG Signals}}

Despite its growing appeal, sEMG signals present several technical challenges, particularly in dynamic real-world human-computer interaction. sEMG signals are naturally non-stationary and often exhibit low signal-to-noise ratios (SNR)~\cite{raez2006techniques}. Surface sEMG recordings are affected by various physiological and environmental factors, including electrical interference, motion artifacts, perspiration, shifting electrode contact, and user fatigue~\cite{tkach2010study, eby2025electromyographic}. Furthermore, sEMG captures aggregate activity from multiple overlapping muscle fibers, which can reduce signal specificity due to anatomical variability and inter-muscle crosstalk~\cite{artemiadis2012emg}. Even minor shifts in electrode placement between sessions can also result in substantial changes in signal patterns, complicating personalization and robust generalization across users.

\paragraph{\textbf{Prior Work in sEMG-to-Text Input}} 

In this work, we focus on sEMG-to-text decoding. Prior research shows that this field has evolved through three main stages: from discrete keypress classification, to sequential gesture mapping, and finally to continuous text decoding from streaming signals. Early studies demonstrated that forearm sEMG signals can clearly distinguish individual keypresses. For example, Sharma et al.~\cite{sharma2017neural} extracted statistical features and used a k-nearest-neighbor classifier guided by reinforcement signals to reliably tell apart finger movements. Fu et al.~\cite{fu2020typing} advanced this work by mapping coordinated three-finger gestures to a T9 layout; their CNN–LSTM model captured both spectral and temporal structures in multi-keypress sequences, moving beyond one-event-at-a-time recognition. Later, Crouch et al.~\cite{crouch2021natural} reframed EMG-based typing as a streaming sequence modeling problem, similar to speech recognition, generating text directly from overlapping muscle activity without explicit segmentation. Building on this idea, Sivakumar et al.~\cite{sivakumar2024emg2qwerty} applied an automatic speech recognition–style model to the \textit{emg2qwerty} dataset, using a time–depth separable CNN (with a 1s receptive field) and band-specific, rotation-invariant filters. They trained the model using Connectionist Temporal Classification (CTC) loss and decoded the outputs with a lexicon-free 6-gram character-level language model, employing modified Kneser–Ney smoothing and beam search. While these systems focus on replicating physical keyboard typing, the next step is to enable virtual keyboard text interaction, which motivates our approach in this work.

\paragraph{\textbf{Open Challenges in the State-of-the-Art}}

Despite recent advancements, sEMG-to-text decoding continues to encounter challenges. sEMG signals are inherently noisy, vary across users, and often convey ambiguous information. Recent findings by Sivakumar et al.~\cite{sivakumar2024emg2qwerty} demonstrated strong within-user accuracy but limited cross-user transfer. Fine-tuned personalized models achieved a 6.9\% character error rate (CER) with a 6-gram language model, while generalized models (cross-user) reached a 51.78\% CER. In the absence of a language model, CER increased to 11.28\% and 55.38\%, respectively, emphasizing the substantial reliance of current systems on language priors. These outcomes highlight the necessity for methods that generalize more effectively across users and leverage language models more efficiently. Future research should focus on developing models that learn transferable muscle-language representations with minimal calibration, directly capture local context from sEMG dynamics, and utilize the language model primarily for long-range support. The ultimate objective is to create on-body sEMG-to-text systems that remain comfortable, fast, and robust for continuous, keyboard-free typing.

\begin{figure}[t]
    \centering
    \includegraphics[width=\linewidth]{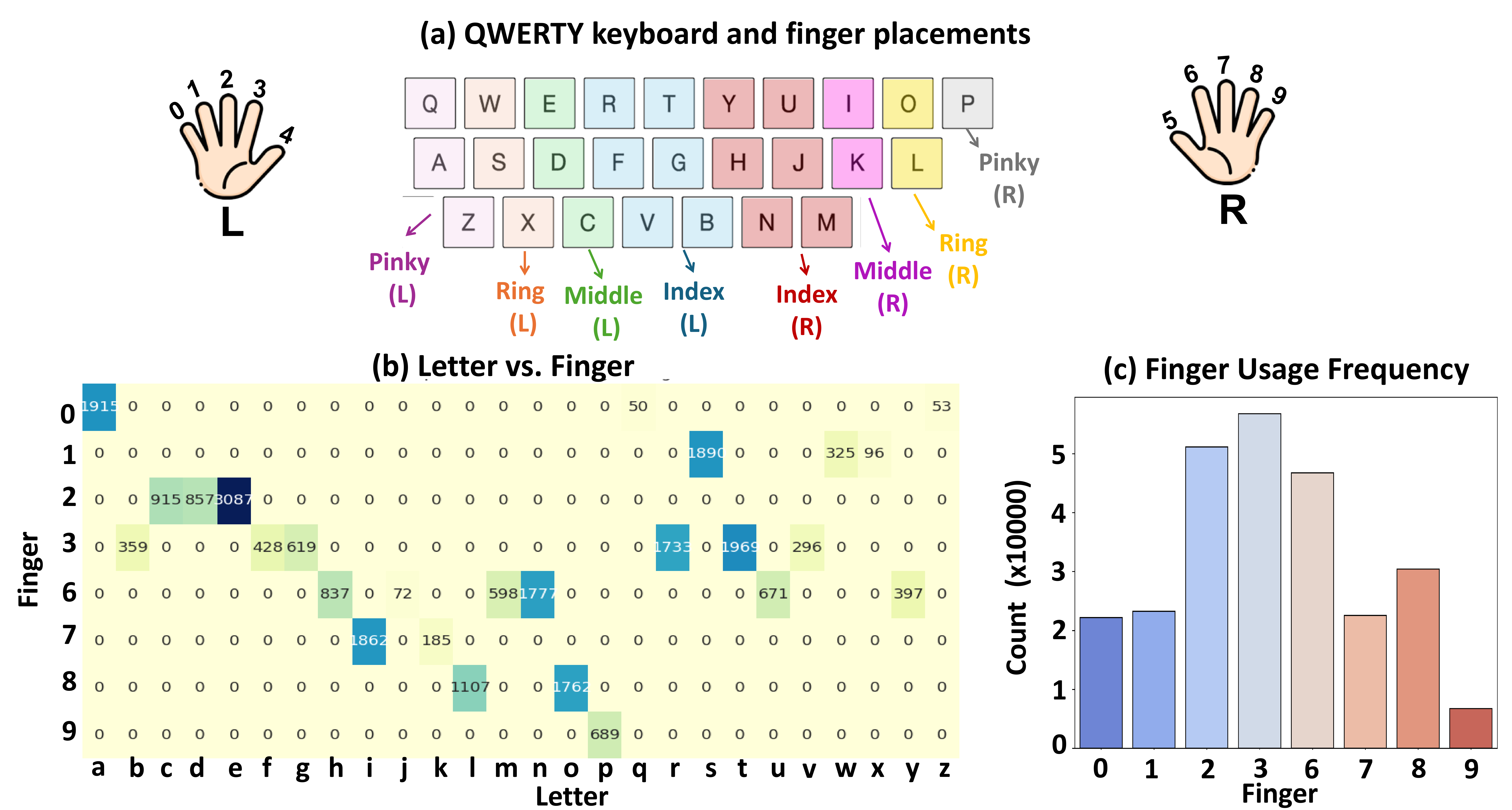}
    \caption{\textbf{QWERTY Ergonomics and Finger Usage.}
(a) The standard QWERTY keyboard layout is depicted, showing the conventional finger-to-key assignments. Letters are distributed among eight active fingers; the thumbs (4, 5) are included for completeness.
(b) Finger-based letter mapping.
(c) Finger usage frequency, highlighting dominance of index and middle fingers in the \textit{emg2qwerty} dataset.}
~\vspace{-4mm}
    \label{fig:char_finger}
\end{figure}

\section{Ergonomic Mapping of Letters to Fingers and the Variability}\label{sec_motivation}

Building towards addressing the above-mentioned challenges and guide our design, we adopt an ergonomic letter-to-finger mapping that captures shared motor patterns and stabilizes decoding across users and sessions. We first infer the active finger from sEMG and then decode the letter within that finger's key set, which reduces the effective search space and supports virtual/keyboard-free operation while maintaining the language model as a lightweight long-range prior.
The QWERTY keyboard layout naturally supports this approach. As the standard for English typing, it balances speed and comfort by minimizing finger movement~\cite{noyes1983qwerty}. As shown in Fig.~\ref{fig:char_finger}(a), touch typing follows an ergonomic rule, each letter key belongs to one of eight fingers (excluding the thumbs)~\cite{TypingLe73:online,Typingwi87:online,WhereSho97:online}. Typists place their fingers on the \textit{home row}, \texttt{ASDF} for the left hand and \texttt{HJKL} for the right, and return there after each keystroke~\cite{Touchtyp90:online}. This habit limits motion, spreads effort evenly, and lets trained typists type faster with less mental effort because they no longer need to decide which finger to use. This mapping reflects long-standing conventions in typing training and is widely used in human-computer interaction and motor-control studies~\cite{zhang2022typeanywhere,zhai2012word,sax2011liquid}. Research has shown that touch typists rely on consistent bimanual finger placement and kinesthetic feedback instead of visual cues~\cite{donica2018keyboarding}, and trained typists build strong muscle-memory patterns that let them type without looking~\cite{gentner1983keystroke}.

We apply the same rule to the \textit{emg2qwerty} dataset by mapping each letter to one of these eight fingers (indexed 0–9, excluding thumbs). In Fig.~\ref{fig:char_finger}(b), we find that common letters such as \texttt{e}, \texttt{t}, \texttt{n}, and \texttt{o} mostly come from the index and middle fingers (2, 3, 6), while rare letters like \texttt{q}, \texttt{z}, and \texttt{p} come from the pinkies (0, 9). This pattern results in structured variation in the sEMG signals, where central fingers move more often and produce steadier signals, while edge fingers activate less often and exhibit greater variability. The finger-usage distribution in Fig.~\ref{fig:char_finger}(c) confirms this asymmetry, reflecting both ergonomic design and meaningful physiological structure. \emph{We also acknowledge that many users deviate from canonical touch-typing (e.g., alternative fingerings, especially in the central keys of the keyboard). To account for this, our evaluation explicitly incorporates typing variability within \textit{emg2qwerty} (\Cref{sec:rq3_fingernoise} assessing performance across heterogeneous typing styles and non-canonical finger–letter assignments.} Decoding models can use this finger-to-letter hierarchy as a built-in structure to reduce noise and generalize better across users and sessions. In short, the finger-based organization, anchored by the home row convention, acts as both a useful constraint and a strong prior for building robust decoders from sEMG signals.

\section{Proposed Methodology: The \SYSTEM{} Framework}

\begin{figure}[t]
    \centering
    \includegraphics[width=0.9\linewidth]{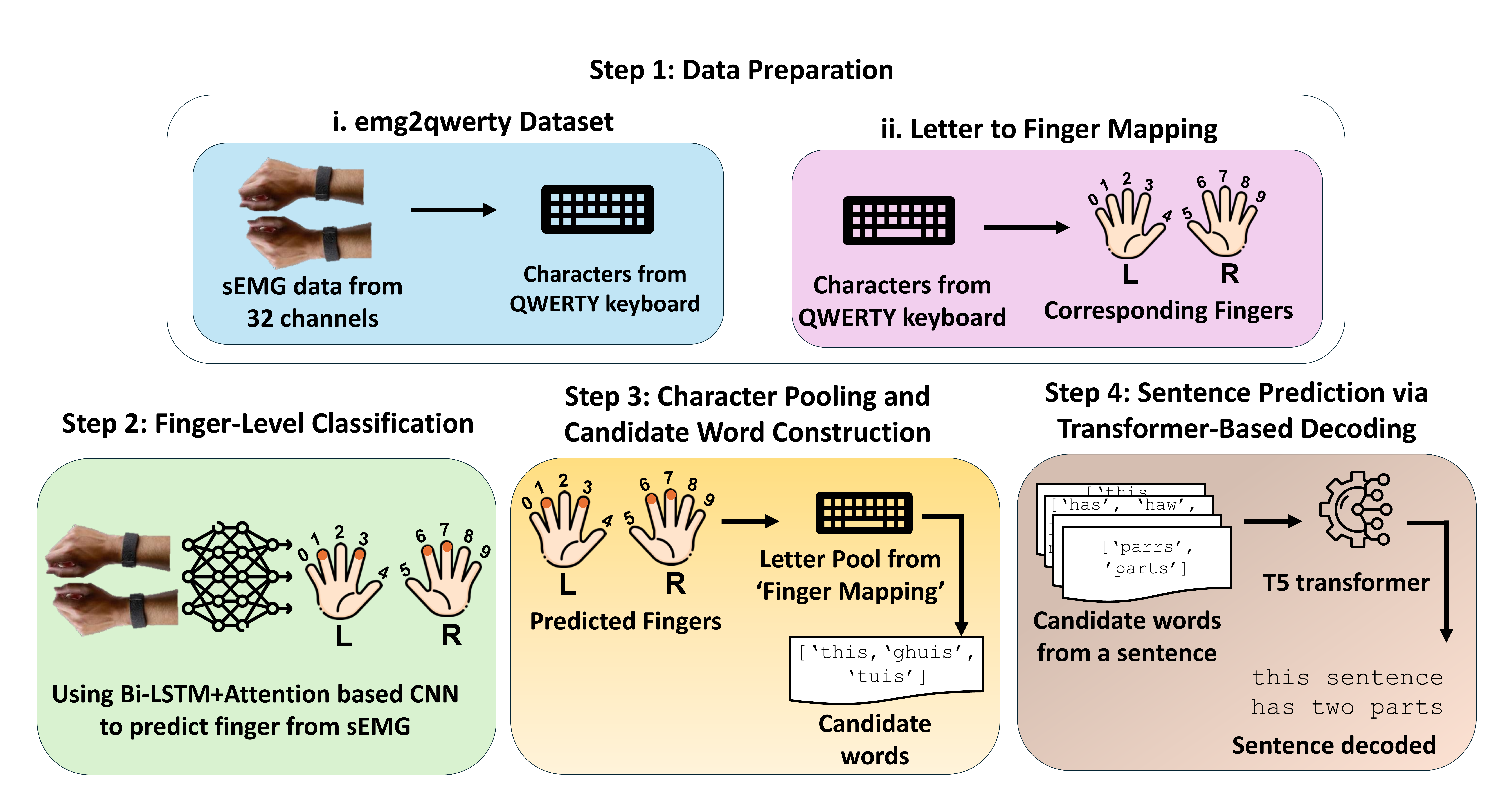}
\caption{\textbf{\SYSTEM{}: Modular sEMG-to-Text Framework.} The four-stage diagram from sEMG signals to sentence decoding: (i) data preparation with ergonomic finger mapping, (ii) finger classification, (iii) letter \& candidate word pooling, and (iv) T5-based decoding.}
    \label{fig:system}
    ~\vspace{-8mm}
\end{figure}

We present \textbf{\SYSTEM{}}, a modular sEMG-to-text decoding framework that mirrors the structure of human typing behavior. As shown in Fig.~\ref{fig:system}, the \SYSTEM{} framework has four steps:  
(i) \textit{Data Preparation}, where we use sEMG signals from \textit{emg2qwerty} dataset and align them with typed letters from a QWERTY keyboard, followed by mapping each letter to its corresponding finger based on typing ergonomics;  
(ii) \textit{Finger-Level Classification}, where a Bi-LSTM with an attention mechanism is used to classify finger activity from the sEMG signals;   
(iii) \textit{Letter Pooling and Candidate Word Construction}, where predicted fingers are mapped to candidate letters and used to generate plausible candidate words;   
(iv) \textit{Sentence Prediction via Transformer-Based Decoding}, where a fine-tuned T5-based transformer language model decodes candidate words into syntactically and semantically plausible sentences. In the following subsections, we detail the design of each step of \SYSTEM{}.

\subsection{Data Preparation}

We start by extracting time-aligned sEMG segments for each keypress from the raw HDF5 logs in the \textit{emg2qwerty} dataset~\cite{GitHubfa40:online}. Although this process is designed for \textit{emg2qwerty}, our modular pipeline can easily adapt to other datasets or preprocessing setups. Each log records synchronized sEMG signals from 32 electrodes (16 on each forearm) along with timestamps for every keystroke. For each keypress, we match the recorded key and its timestamp to locate the corresponding sEMG interval, and then extract a short segment that includes 1s before and after the event to capture both preparatory and residual muscle activity. We keep all 32 channels at their original resolution and save the processed segments as compressed files to speed up later analysis.

Next, we label each letter with the finger that typically presses it, following standard typing ergonomics~\cite{TypingLe73:online, Typingwi87:online, WhereSho97:online} and the QWERTY keyboard layout~\cite{noyes1983qwerty}. We map letters to eight active fingers (indices 0–3 for the left hand and 6–9 for the right) while reserving the thumbs for the spacebar. \Cref{tab:key-finger-split} shows the full key-to-finger mapping. Although not every user follows the exact QWERTY convention, prior studies in typing biomechanics and sEMG research have used this mapping as a consistent baseline. Classic ergonomics work measured fine-wire sEMG signals from ten expert typists~\cite{fernstrom1994electromyographic}, and later studies continued to recruit experienced typists for controlled experiments~\cite{simoneau2003effect,eby2025electromyographic,wang2024efficient}. The \textit{emg2qwerty} dataset~\cite{sivakumar2024emg2qwerty} also screened participants to ensure they could type without looking at the keyboard and follow the correct finger-to-key mapping at least 90\% of the time. This consistency supports the continued use of standardized mappings in sEMG-based typing research. To verify how well our method handles user variation, we also added controlled perturbations to the finger-to-key mapping during evaluation to test robustness (see~\Cref{sec:rq3_fingernoise}). The results show that even when users deviate from the standard mapping, our method maintains strong performance, confirming its robustness to individual typing styles.



\subsection{Finger-Level Classification}  
\label{sec:finger-classification}

Given the extracted (sEMG segment, finger label) pairs, we train a model that predicts which finger generates each muscle signal directly from the raw multichannel sEMG input. By framing the task at the finger level instead of the letter level, we reduce the classification space from 26 letters to 8 fingers, thereby better matching how real muscle activations occur during typing. This design helps the model handle problems such as letter-level class imbalance and overlapping muscle activity between letters, making it more robust and easier to generalize to new users.

\begin{figure}[t]
    \centering
    \begin{minipage}{0.55\linewidth}
        \centering
        \captionof{table}{\textbf{Letter to Finger Mapping}}
        \label{tab:key-finger-split}
        \includegraphics[width=\linewidth]{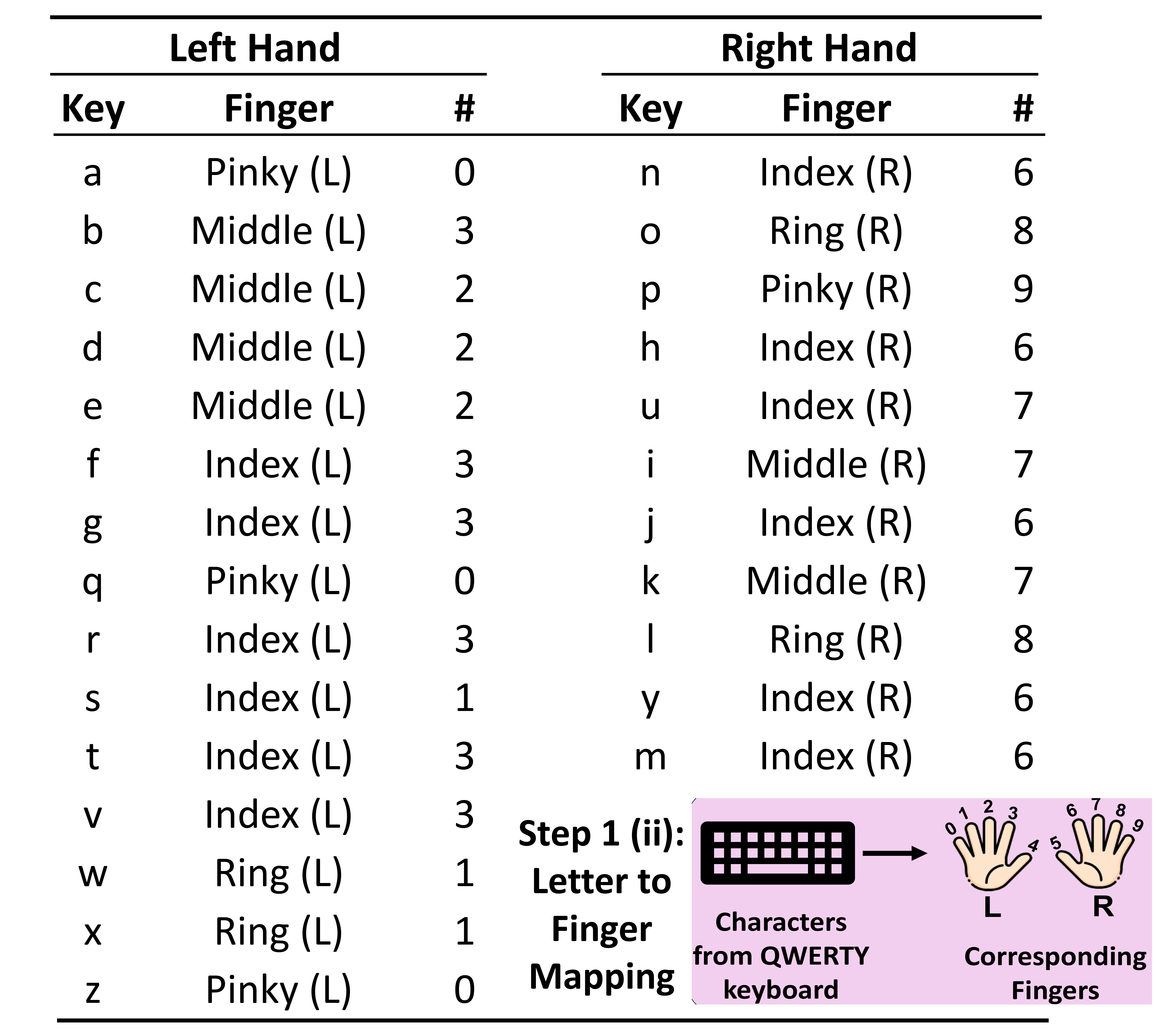}
    \end{minipage}
    \hfill
    \begin{minipage}{0.42\linewidth}
        \centering
        \includegraphics[width=\linewidth]{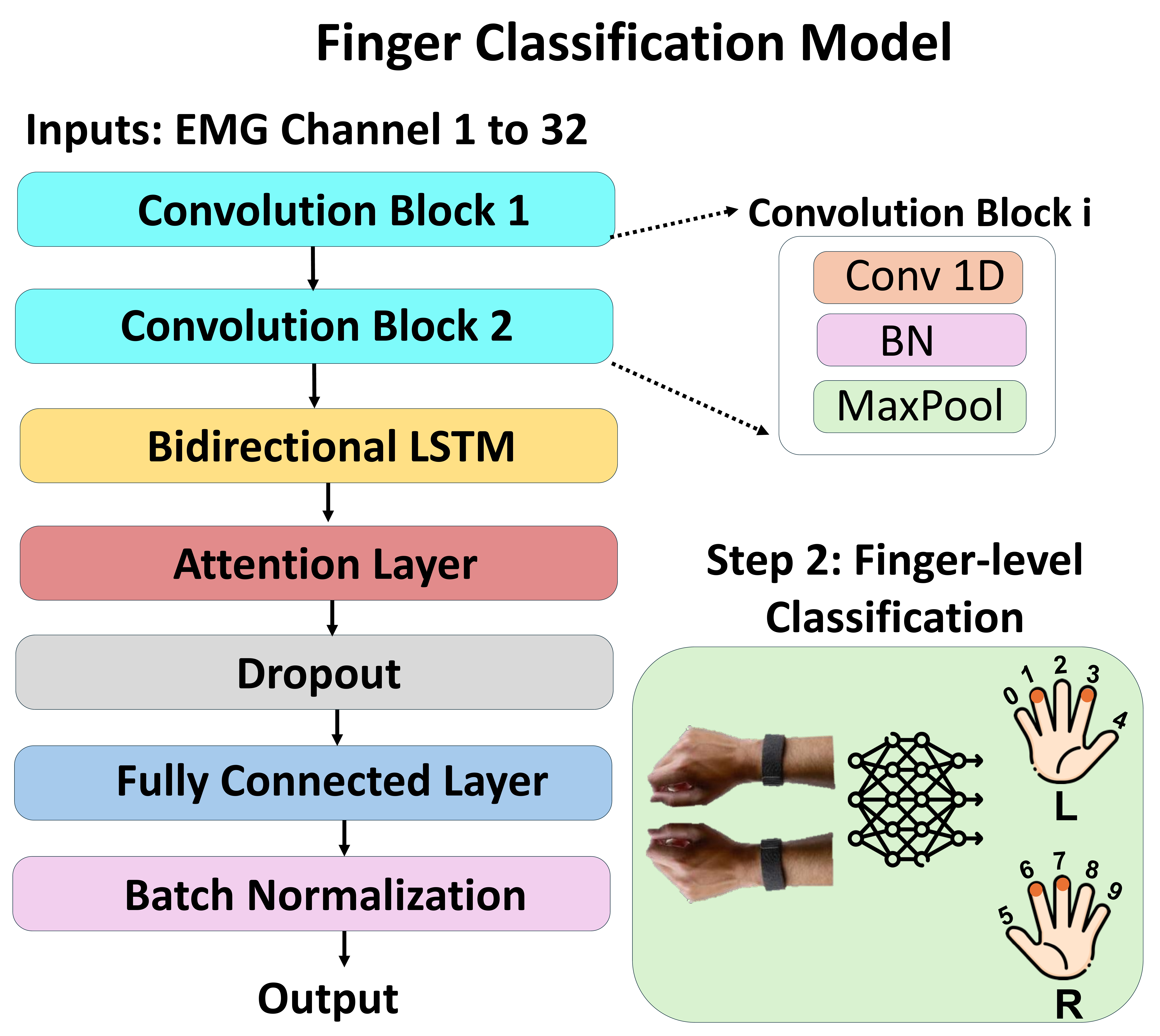}
        \caption{\textbf{Finger Classification Network.} CNN–BiLSTM–Attention architecture for predicting finger labels from multichannel sEMG.}
        \label{fig:finger_classification}
    \end{minipage}
\end{figure}

The finger-level classification is a supervised learning problem with two goals: (i) capture short-term correlations across multiple sEMG channels, and (ii) model bidirectional temporal dependencies, since muscle activations often include both anticipatory and residual signals. We use CNNs to learn local spatiotemporal features and LSTMs to track longer dependencies in both time directions~\cite{fan2024bilstm,sainath2015convolutional,li2023medical}. Building on these strengths, our model stacks convolutional blocks followed by a bidirectional LSTM layer. We also incorporate an attention mechanism to highlight the most informative time steps, and fully connected layers refine the representation to predict the correct finger class. In short, our architecture combines CNN, BiLSTM, and attention modules with fully connected layers for final classification (Fig.~\ref{fig:finger_classification}).

The model takes an input sequence $\mathbf{X}\in\mathbb{R}^{C\times T}$, where $C$ is the number of electrode channels and $T$ is the number of time steps, and outputs the predicted finger label $y \in \{0,\ldots,7\}$. The architecture runs through four stages in sequence:

\begin{itemize}
\item \textbf{Convolutional feature extraction.}  
A convolutional layer applies $\mathbf{H}^{l}=\sigma(\mathbf{W}^{l} * \mathbf{H}^{l-1}+\mathbf{b}^{l})$ with $\mathbf{H}^{0}=\mathbf{X}$, where $*$ denotes the convolution operation, $\sigma$ is the ReLU activation, and $\mathbf{W}^{l},\mathbf{b}^{l}$ are trainable kernels and biases. Each convolutional filter learns to detect localized temporal features in the sEMG waveform. Max-pooling reduces the temporal dimension from $T$ to $T'$. This stage thus encodes short-term dependencies and frequency-specific energy variations related to motor unit recruitment.

\item \textbf{Temporal modeling with BiLSTM.}  
The convolved features $\mathbf{H}^{L}\in\mathbb{R}^{F\times T'}$ (where $F$ is the number of learned feature maps) are passed to a bidirectional LSTM to model long-range temporal dependencies. The forward update $\overrightarrow{\mathbf{h}}_{t}=\mathrm{LSTM}_f(\overrightarrow{\mathbf{h}}_{t-1},\mathbf{H}^{L}_{t})$ captures past context, while the backward update $\overleftarrow{\mathbf{h}}_{t}=\mathrm{LSTM}_b(\overleftarrow{\mathbf{h}}_{t+1},\mathbf{H}^{L}_{t})$ captures future context. The concatenated hidden state $\mathbf{h}_t=[\overrightarrow{\mathbf{h}}_{t};\overleftarrow{\mathbf{h}}_{t}]$ encodes bidirectional temporal information, preserving both anticipatory and residual activity. This is crucial for sEMG, where muscle activation and relaxation phases often overlap, and motion intent can be inferred from both preceding and succeeding signal dynamics.

\item \textbf{Attention-based focus.}  
To emphasize physiologically informative segments, the model learns an attention distribution over time. Each BiLSTM output $\mathbf{h}_t\in\mathbb{R}^{d_h}$ represents a $d_h$-dimensional hidden feature vector capturing the temporal EMG context at time $t$, including contraction intensity and transition dynamics. The model computes a relevance score $e_t=\mathbf{v}^{\top}\tanh(\mathbf{W}_a\mathbf{h}_t)$, where $\mathbf{W}_a\in\mathbb{R}^{d_a\times d_h}$ is a trainable projection matrix mapping hidden states into an attention space, and $\mathbf{v}\in\mathbb{R}^{d_a}$ is a context vector that measures their physiological relevance. The normalized weight $\alpha_t=\exp(e_t)/\sum_{t'}\exp(e_{t'})$ quantifies the contribution each time, and the context vector $\mathbf{c}=\sum_t\alpha_t\mathbf{h}_t$ aggregates features proportionally to their learned importance. 

\item \textbf{Fully connected layers.}  
The context vector $\mathbf{c}\in\mathbb{R}^{d_c}$ is passed through a dense layer with ReLU to compute $\mathbf{z}=\phi(\mathbf{W}_f\mathbf{c}+\mathbf{b}_f)$, followed by dropout and batch normalization. The classifier output is $\hat{\mathbf{y}}=\mathrm{softmax}(\mathbf{W}_o\mathbf{z}+\mathbf{b}_o)$ with $y\in\{0,\ldots,7\}$, $\hat{y}_k\in[0,1]$, and $\sum_{k=0}^{7}\hat{y}_k=1$; for $K=8$ classes, the cross-entropy loss is $\mathcal{L}=-\sum_{k=0}^{K-1}\mathbb{1}[y=k]\log\hat{y}_k=-\log\hat{y}_y$. 
\textit{Where:} $\phi$ is ReLU; $\mathbf{z}\in\mathbb{R}^{d_z}$; $\mathbf{W}_f\in\mathbb{R}^{d_z\times d_c}$, $\mathbf{b}_f\in\mathbb{R}^{d_z}$; $\hat{\mathbf{y}}\in\mathbb{R}^{K}$ with $K=8$; $\mathbf{W}_o\in\mathbb{R}^{K\times d_z}$, $\mathbf{b}_o\in\mathbb{R}^{K}$; $y$ is the ground-truth class index; $\hat{y}_k$ is the predicted probability for class $k$; $\mathbb{1}[y=k]$ is the indicator function. \textit{Note: if $\mathbf{c}=\sum_t\alpha_t\mathbf{h}_t$ directly, then $d_c=d_h$.}

\end{itemize}

This model combines convolutional blocks, BiLSTM units, attention, and fully connected layers to blend local spatial features with temporal and contextual information. It aligns the classification task with how people type, grounding the architecture in real muscle and finger movements. As a result, the model can learn multichannel sEMG signals and predict which finger is active. These finger predictions then feed into the next stage of the \SYSTEM{} pipeline, where the system maps fingers to possible letters and begins reconstructing words.



\begin{figure}[t]
    \centering
    \captionof{table}{\textbf{Examples of Finger-to-Letter Mapping and Word Candidate Pooling.}}
    \vspace{-2mm}
    \includegraphics[width=1\linewidth]{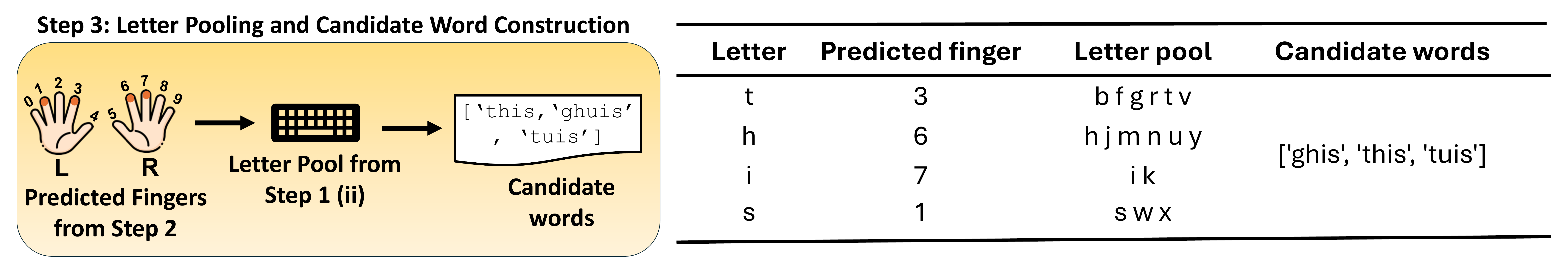}
    \label{tab:finger-letter-candidates}
        \vspace{-4mm}
\end{figure}

\subsection{Letter Pooling and Candidate Word Construction} 
\label{sec:char-pooling}

After predicting which finger produced each keystroke, we restrict the letter search space using a fixed mapping from fingers to letter subsets based on the standard QWERTY touch-typing layout. \Cref{tab:finger-letter-candidates} lists some examples of the finger-to-letter mapping. For instance, finger index three maps to letters including \texttt{b}, \texttt{f}, \texttt{g}, \texttt{r}, \texttt{t}, and \texttt{v}, which people normally type with the left index finger. This step both reduces the decoding search space and adds biomechanical knowledge to the prediction process, making the model faster and more realistic. For each predicted finger sequence for a word (e.g., \textit{this} $\rightarrow$ [3, 6, 7, 1]), we map each finger to its corresponding set of possible letters to produce a \emph{letter pool sequence} that lists possible letters for each position in the word.

From this letter pool sequence, the next step is to build \emph{candidate word pool}. The ideafirst to take the Cartesian product of all letter poolsools to generate every possible word combination, and then filter these words using large English corpora such as the Brown Corpus~\cite{NLTKSamp10:online}, WordNet~\cite{pedersen2004wordnet}, and public English dictionaries~\cite{englishw41:online}. This filtering step removes invalid or nonsensical combinations (e.g., \emph{``bbs''}, \emph{``jnd''}) while preserving real English words. \Cref{tab:word-reconstruction} shows examples of the resulting candidate word pools.


\begin{table*}[t]
  \caption{\textbf{Examples of Candidate Word Pools Using Finger-Based Letter Pools.}}
  \small
  \label{tab:word-reconstruction}
  \begin{tabular}{L{1.8cm} L{7.1cm} L{5.4cm}}
    \toprule
    \textbf{Sentence} & \textbf{Letter Pool Sequence} & \textbf{Candidate Words} \\
    \midrule
    this  & [b f g r t v ; h j m n u y ; i k ; s w x] & ghis, this, tuis \\
    has   & [h j m n u y ; a q z ; s w x]             & has, haw, jaw, mas, max, naw, yas, yaw \\
    two   & [b f g r t v ; s w x ; l o]               & two \\
    parts & [p ; a q z ; b f g r t v ; b f g r t v ; s w x] & parrs, parts \\
    \midrule
    legitimate & [l o ; c d e ; b f g r t v ; i k ; b f g r t v ; i k ; h j m n u y ; a q z ; b f g r t v ; c d e] & legitimate \\
    load       & [l o ; l o ; a q z ; c d e]           & load, ooze \\
    economic   & [c d e ; c d e ; l o ; h j m n u y ; l o ; h j m n u y ; i k ; c d e] & economic \\
    women      & [s w x ; l o ; h j m n u y ; c d e ; h j m n u y] & soncy, women, wouch \\
    \bottomrule
  \end{tabular}
\end{table*}



\subsection{Sentence Prediction via Transformer-Based Decoding}
\label{sec:sent-decoder}

In the final stage of our framework, we build complete sentences from the candidate word pools. The first-stage finger classification introduces some level of uncertainty, which can propagate to the word pools. To handle this, the sentence decoder should examine the entire sentence instead of focusing only on nearby words. Therefore, we utilize a transformer-based language model that processes all candidate sets simultaneously and selects one word from each set to form the final, coherent sentence.

\paragraph{\textbf{T5 Transformer.}} We use the Text-to-Text Transfer Transformer (T5) introduced by Google in 2019~\cite{raffel2020exploring}. T5 was pretrained on the Colossal Clean Crawled Corpus (C4) with a vocabulary size $|\mathcal{V}|\!\approx\!32,000$. The T5 model includes an encoder and a decoder. The encoder has multiple layers (e.g., 12 in T5-Base), each with bidirectional self-attention, a two-layer feed-forward network with ReLU activation, layer normalization, and residual connections for stable training. The decoder mirrors this structure but adds masked self-attention to avoid looking at future tokens and includes encoder–decoder attention for contextual generation~\cite{raffel2020exploring}. T5 is a general sequence-to-sequence model that can be fine-tuned for tasks such as summarization, question answering, and text classification~\cite{atitienei2024exploring,awalurahman2024automatic,marshan2024medt5sql,zhao2025leveraging,ling2024automatic,usui2023translation}. Its scalability and strong performance on benchmarks such as GLUE and SuperGLUE~\cite{wang2018glue,wang2019superglue} have made it widely used in downstream applications.

We choose T5 for our sentence decoding task because its encoder–decoder structure enables global reasoning across the entire sentence. The encoder captures bidirectional context from the input sequence, while the autoregressive decoder generates text by conditioning on both the encoder outputs and previously generated tokens. This design lets the model use later context to fix earlier uncertainties. For example, distinguishing ``has'' from ``haw'' when it processes the phrase ``two parts'', which is essential for accurate decoding in \SYSTEM{}.

\begin{figure}[t]
    \centering
    \captionof{table}{\textbf{Examples of Using Candidate Word Pools to Generate Sentences.}}
    \includegraphics[width=0.85\linewidth]{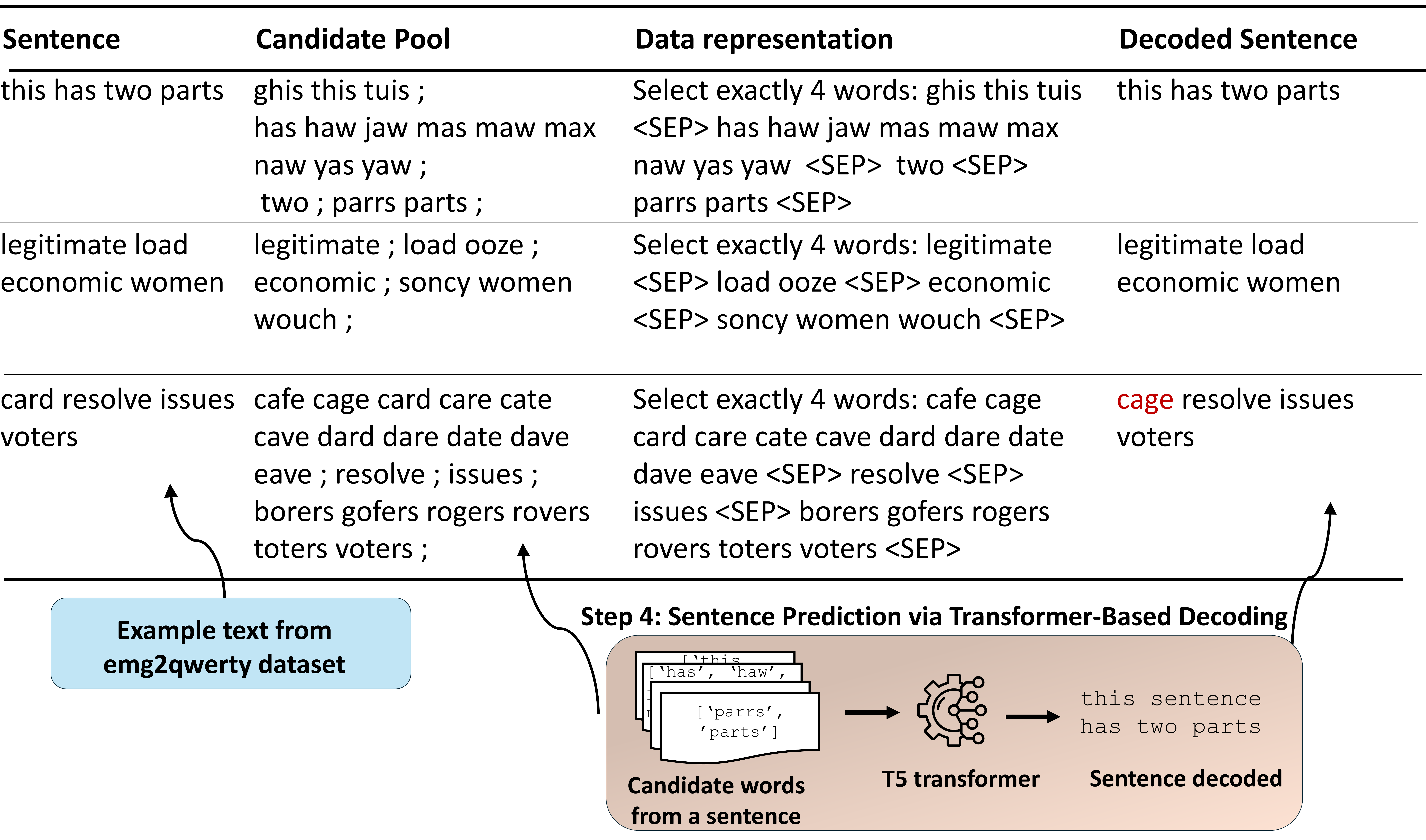}
    \label{tab:sentence-reconstruction}
\end{figure}

\paragraph{\textbf{Input Representation Adaptation.}}  

Unlike typical sequence-to-sequence tasks where the input is already a token sequence, \SYSTEM{}'s sentence decoder takes structured candidate word pools (after Step 3). Standard T5 cannot directly process such structured inputs because it expects a flat text sequence. To address this, we turn a sequence of candidate word pools $\mathcal{G}_1,\ldots,\mathcal{G}_N$ (where $N$ is the sentence length) into an instruction-based prompt of string using a delimiter token \texttt{<SEP>} for separation, i.e., ``Select exactly $N$ words: $\mathcal{G}_1$ \texttt{<SEP>} $\mathcal{G}_2$ \texttt{<SEP>} \ldots $\mathcal{G}_N$ \texttt{<SEP>}''. This constrains that at position $n$, \SYSTEM{} only allows choices from the corresponding candidate word pool. For example, in column 3 of~\Cref{tab:sentence-reconstruction}, the sentence \emph{``this has two parts''} with candidate word pools $\mathcal{G}_1=\{\text{ghis},\text{this},\text{tuis}\}$, $\mathcal{G}_2=\{\text{has},\text{haw}\}$, $\mathcal{G}_3=\{\text{two}\}$, $\mathcal{G}_4=\{\text{parrs},\text{parts}\}$ produces the serialized input $S=\phi(\mathcal{G}_1,\ldots,\mathcal{G}_4)$: ``Select exactly 4 words: ghis this tuis \texttt{<SEP>} has haw \texttt{<SEP>} two \texttt{<SEP>} parrs parts \texttt{<SEP>}''. 

\paragraph{\textbf{Sentence Generation via Constrained Beam Search.}}

We integrate beam search with a fine-tuned T5 model to generate sentences from the candidate word pools. T5's encoder reads the adapted serialized input string once and produces contextual embeddings that represent all possible words jointly. During decoding, T5 generates the sentence word by word. At each position n, T5 only considers words from the corresponding candidate pool $\mathcal{G}_n$ as valid options. For each candidate word, we compute its probability under the T5 decoder, conditioned on both the encoder context and the words already chosen. Beam search then maintains a set of the top-$B$ (where $B$ is the beam width) partial sentences, expands each one with all candidate words in $\mathcal{G}_n$, and retains the highest-scoring continuations. After all positions are filled, \SYSTEM{} selects the sentence with the best total score. This process enables T5 to apply its global language knowledge from pretraining while beam search systematically explores and scores multiple word combinations, producing a final sentence that is both fluent and consistent with the physiological decoding constraints. Column 4 of~\Cref{tab:sentence-reconstruction} lists examples of the decoded sentences from candidate word pools.

\section{Evaluation}

In this section, we evaluate \SYSTEM{}. We report finger-level classification accuracy, sentence-level error (CER/WER), robustness to typing variability and cross-dataset shift, and end-to-end comparisons with prior sEMG-to-text systems. We also analyze the \textit{emg2qwerty} dataset to assess signal quality. Unless otherwise noted, all experiments use the \textit{emg2qwerty} dataset. We address the following research questions (RQs):

\begin{itemize}
  \item \textbf{RQ1:} What is the performance of \SYSTEM{}'s modular decoding framework, from finger-level classification to sentence generation?
  \item \textbf{RQ2:} How well does \SYSTEM{} generalize to longer, naturalistic text and non-canonical typing patterns?
  \item \textbf{RQ3:} Does the \textit{emg2qwerty} dataset provide sufficient scale and keystroke-level alignment to serve as a reliable benchmark for sEMG-to-text decoding?
  \item \textbf{RQ4:} How does \SYSTEM{} position itself within the evolution of sEMG-to-text systems, and what insights does it provide for deployment?
\end{itemize}


\subsection{RQ1: Performace of \SYSTEM{}}
\label{sec:rq1}

We evaluated \SYSTEM{} using the \textit{emg2qwerty} dataset because it is the most diverse sEMG-to-text dataset available released in 2024~\cite{sivakumar2024emg2qwerty}. It consists of synchronized sEMG recordings from both forearms (16 channels per side) along with annotated keystroke events. We extracted structured examples containing the user ID, keystroke timing, and the multi-channel sEMG segments. This ensures that every keystroke is paired with its corresponding sEMG window. The data preprocessing pipeline was run on Google Colab using CPU resources and took about 15 hours to complete. In total, we evaluated using data from 30 participants. For finger classification, we trained the neural network model with the Adam optimizer~\cite{kingma2014adam} (learning rate \(10^{-3}\), batch size 32) for 100 epochs. To prevent overfitting, we applied a combination of regularization strategies, including an adaptive learning rate scheduler (ReduceLROnPlateau~\cite{ReduceLR7:online} with a decay factor of 0.01 when the validation loss plateaued) and early stopping with a patience of 10 epochs. For sentence decoding, we fine-tuned a pretrained T5 transformer~\cite{raffel2020exploring} using teacher forcing. Fine-tuning was performed with a batch size of 32, a learning rate of \(5 \times 10^{-5}\) for up to 30 epochs using the Hugging Face Trainer API. All models were implemented in \texttt{TensorFlow}, with pretrained models imported from \texttt{Hugging Face}~\cite{T511:online}. Training and fine-tuning were conducted on either an NVIDIA T4 or an  NVIDIA L40, depending on availability on Google Colab.

\subsubsection{\textbf{Finger-Level Classification}}

A key design choice in \SYSTEM{} is whether to classify finger activations (8 active fingers) first or directly predict letters (26 classes) as the final output. An RNN-based approach is a natural fit for this problem, as recurrent models such as LSTMs are widely used for time-series decoding and can effectively capture temporal dependencies in sEMG signals. Our preliminary experiments showed that averaging left and right arm signals into two input channels significantly degraded performance (40.3\% accuracy) due to the loss of spatial detail. In contrast, using the full 32-channel input in the same CNN–LSTM architecture improved accuracy to 71.2\%, underscoring the importance of high-resolution sEMG representation. Hence, we trained two variants of the same CNN–BiLSTM–Attention architecture (\Cref{sec:finger-classification}), differing only in their output targets:

\begin{itemize}
    \item Model 1 (Finger Classification): Predicts 8 finger labels; achieves 85.4\% average accuracy.
    \item Model 2 (Letter Classification): Predicts 26 letters directly; achieves 51.2\% average accuracy.
\end{itemize}

\begin{wraptable}[12]{r}{0.6\textwidth}
\centering
\vspace{-1mm}
\caption{\textbf{Finger vs. Letter Variant Accuracy (BiLSTM+Attn).}}
\vspace{-1mm}
\label{tab:kfold-accuracy}
\setlength{\tabcolsep}{3pt}
\renewcommand{\arraystretch}{1.05}
\small
\begin{tabular}{lcccc}
\toprule
\multirow{2}{*}{\textbf{Fold}} & \multicolumn{2}{c}{\textbf{Model 1 (\%)}} & \multicolumn{2}{c}{\textbf{Model 2 (\%)}} \\
\cmidrule(lr){2-3} \cmidrule(lr){4-5}
& \textbf{CV} & \textbf{LOUO} & \textbf{CV} & \textbf{LOUO} \\
\midrule
1 & 82.71 & 85.52 & 43.85 & 29.76 \\
2 & 82.01 & 85.41 & 47.52 & 28.46 \\
3 & 88.66 & 80.98 & 51.2 & 30.84 \\
4 & 89.30 & 78.84 & 54.87 &  25.52\\
5 & 84.61 & 77.27 & 58.55 & 26.73 \\
\midrule
\textbf{Mean $\pm$ Std} & \textbf{85.45 $\pm$ 3.36} & \textbf{81.4 $\pm$ 3.76} &
\textbf{51.2 $\pm$ 5.81} & \textbf{28.5 $\pm$ 2.25} \\
\bottomrule
\end{tabular}
\end{wraptable}

\begin{figure}[t]
    \centering
    \includegraphics[width=1\linewidth]{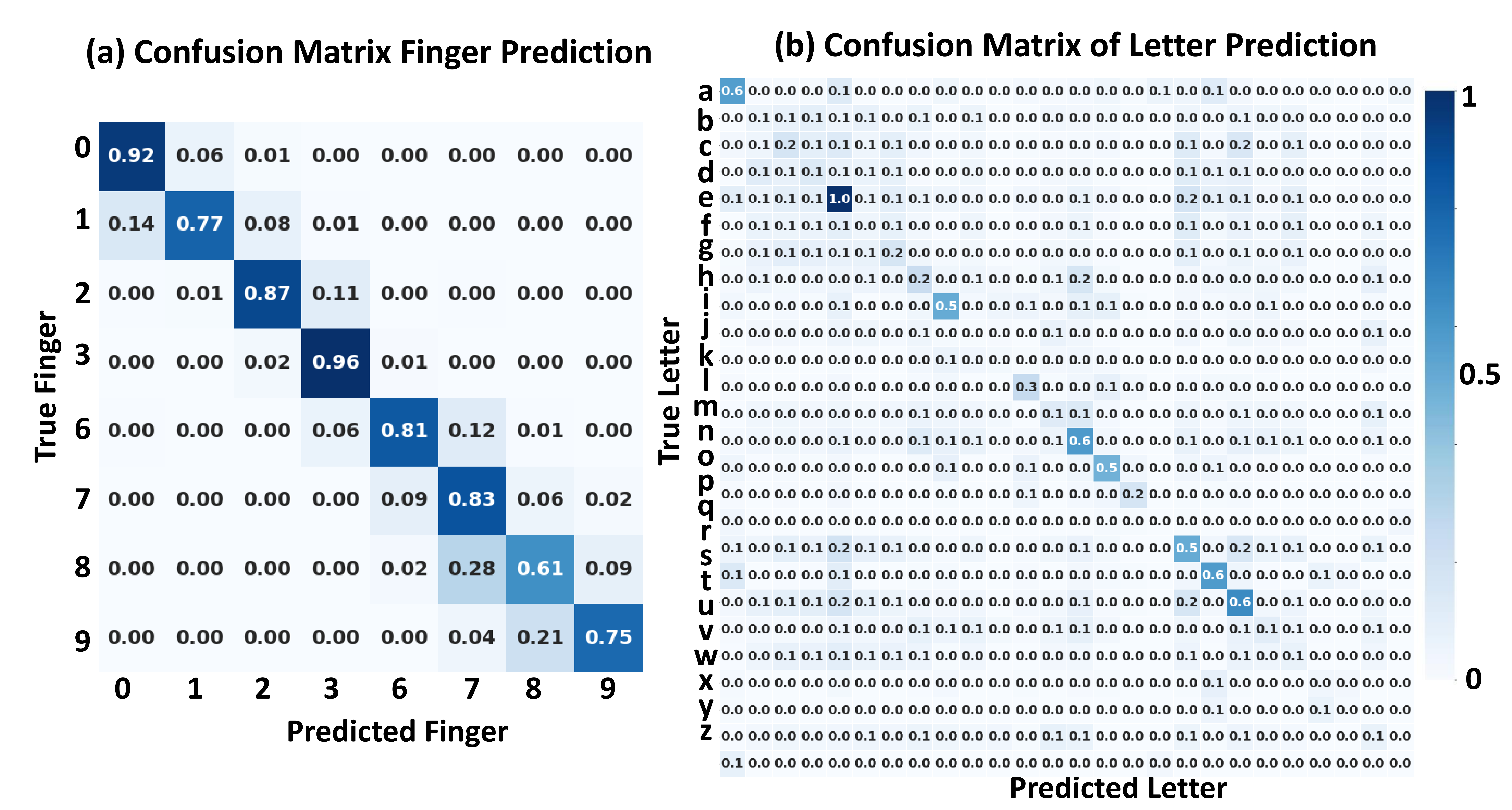}
    \caption{\textbf{Confusion matrices: Finger vs. Letter Prediction} (a) Model 1 (b) Model 2}
    \label{fig:CM}
\end{figure}

This performance gap highlights the tractability of finger-level decoding. sEMG signals are activated due to finger activity, and hence tend to cluster by finger activation rather than by individual letters, since multiple keys share similar muscular patterns when typed by the same finger. This overlap makes letter-level classification significantly more ambiguous. Fig.~\ref{fig:CM} illustrates this contrast. The confusion matrix for Model 1 (Fig.~\ref{fig:CM}a) shows clear diagonal dominance, indicating clean finger separation. In contrast, Model 2’s confusion matrix (Fig.~\ref{fig:CM}b) reveals substantial misclassifications, particularly among spatially adjacent keys (or those involving similar muscular activations). We further evaluated the models' robustness using two validation protocols. As shown in Table~\ref{tab:kfold-accuracy}, 5-fold cross-validation yields an average accuracy of 85.45\% $\pm$ 3.36\%, while leave-one-user-out (LOUO) validation on five held-out users achieves 81.4\% $\pm$ 3.76\% on Model 1. The results of Model 2 are shown in the same table. \textit{While our models follow the standard QWERTY finger-to-key mapping (Table~\ref{tab:key-finger-split}), there may be user-specific deviations like typing `B' with the right index finger instead of the left. We consider such scenarios in~\Cref{sec:rq3_fingernoise}, and our downstream transformer decoder effectively resolves these inconsistencies using sentence-level context, ensuring robustness to both finger prediction errors and real-world typing variability.}

\subsubsection{\textbf{Sentence-Level Prediction}}

Following finger classification, we decode candidate word pools (\Cref{sec:char-pooling}) into complete sentences using a fine-tuned T5 transformer~\cite{raffel2020exploring}. Sentence reconstruction performance is evaluated using Word Error Rate (WER) and Character Error Rate (CER), defined in Equations~\ref{eq:wer} and~\ref{eq:cer}, where $S$, $D$, and $I$ represent substitutions, deletions, and insertions, and $N$ is the total number of reference words or characters.

\begin{figure}[h]
\centering
\begin{minipage}[t]{0.45\linewidth}
\begin{equation}
\text{WER} = \frac{S_W + D_W + I_W}{N_W}
\label{eq:wer}
\end{equation}
\end{minipage}
\hfill
\begin{minipage}[t]{0.45\linewidth}
\begin{equation}
\text{CER} = \frac{S_C + D_C + I_C}{N_C}
\label{eq:cer}
\end{equation}
\end{minipage}
\end{figure}

\paragraph{\textbf{Model Variants and Results}}

We evaluate three T5-base variants, each trained on the candidate-to-sentence generation task:
\begin{itemize}
    \item \textbf{T5 Model 1:} Baseline with standard cross-entropy loss ($\mathcal{L}_{\text{M1}} = \mathcal{L}_{CE}$).
    \item \textbf{T5 Model 2:} Adds a substitution-aware auxiliary loss ($\mathcal{L}_{\text{M2}} = \mathcal{L}_{CE} + \lambda \mathcal{L}_{aux}$).
    \item \textbf{T5 Model 3:} Incorporates Model 2's loss, plus weight decay, label smoothing, and early stopping.
\end{itemize}

\begin{wrapfigure}[14]{r}{0.45\textwidth}
\centering
\vspace{-8mm}
\includegraphics[width=1\linewidth,height=5cm]{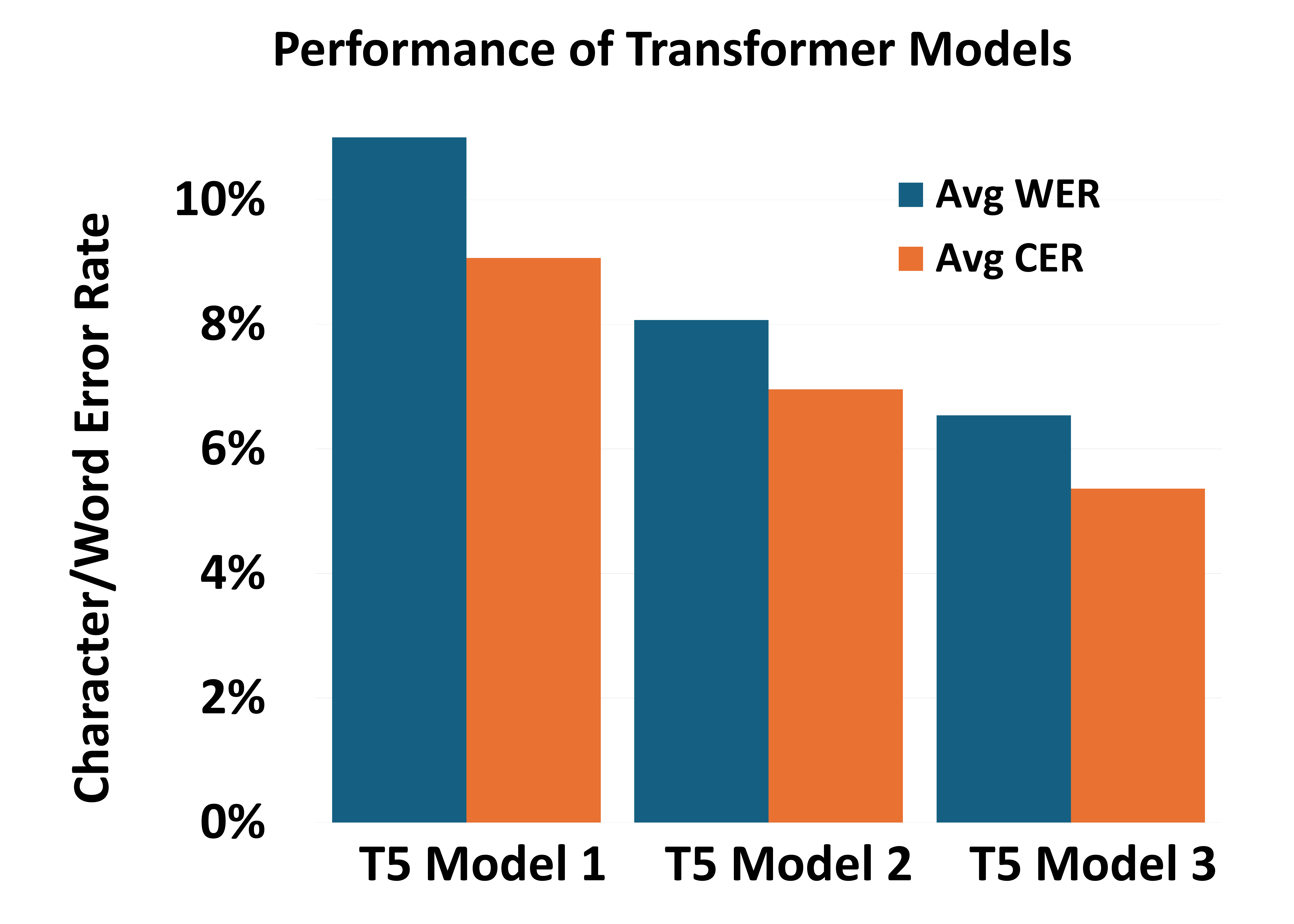}
\vspace{-9mm}
\caption{\textbf{T5 Model Performance.} CER and WER across T5 variants. Lower is better.}
\label{fig:t5_perf}
\end{wrapfigure}

Fig.~\ref{fig:t5_perf} shows that T5 Model 3 achieves the best performance, with 6.5\% WER and 5.4\% CER. These results indicate that the decoder effectively leverages sentence-level context to overcome noise from finger-level predictions and candidate ambiguity.

\paragraph{\textbf{Error Analysis}}

\begin{figure}[h]
\centering
    \captionof{table}{\textbf{Comparisons between Reference and Predicted Sentences in the emg2qwerty dataset.}}
    ~\vspace{-3mm}
    \includegraphics[width=0.9\linewidth]{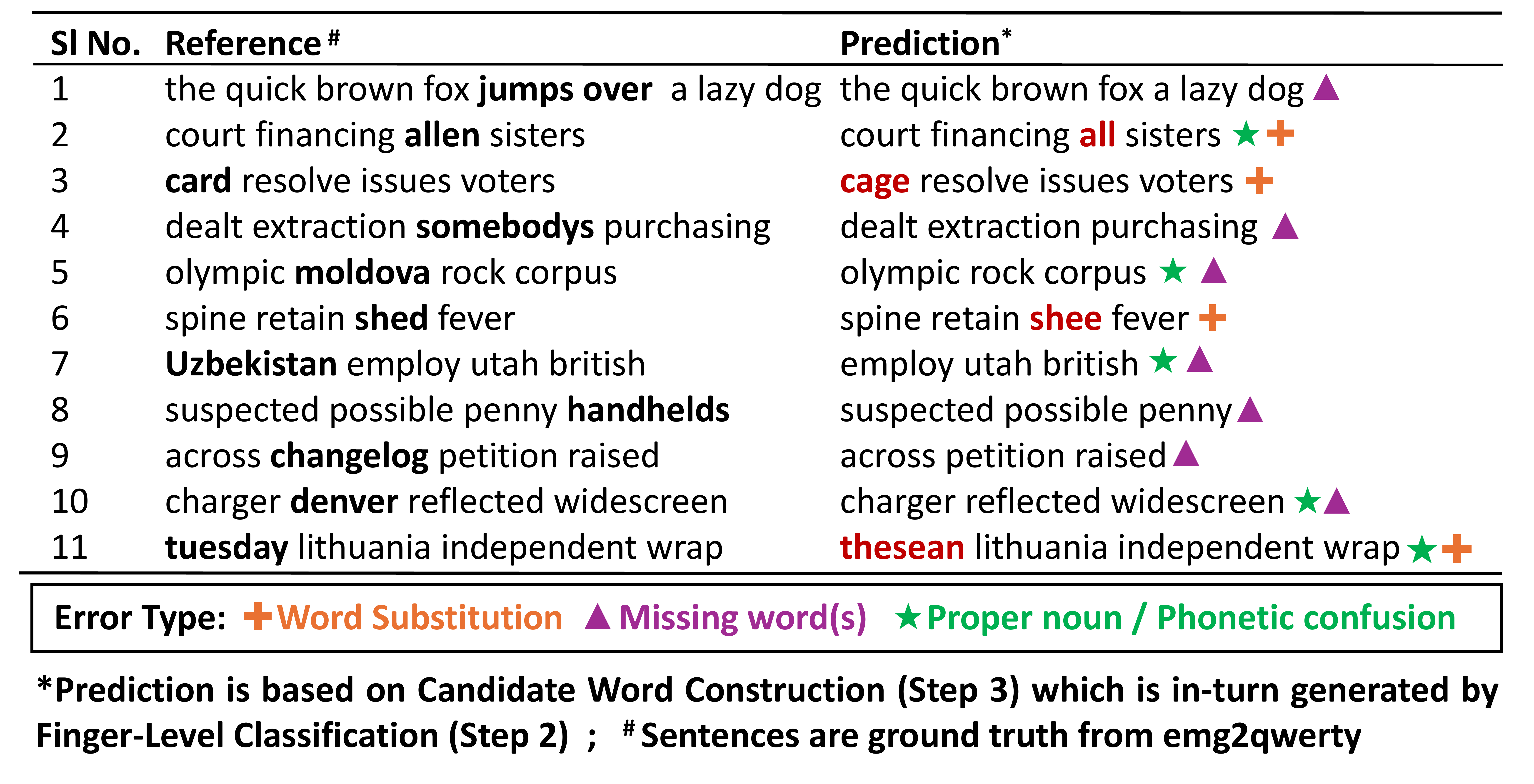}
    \vspace{-4mm}
    \label{tab:ref_pred_errors}
\end{figure}

Table~\ref{tab:ref_pred_errors} presents representative decoding errors. Most fall into three categories:  
(i) semantically or phonetically similar substitutions (e.g., ``card'' $\rightarrow$ ``cage''),  
(ii) omitted proper nouns (e.g., ``Uzbekistan'', ``Moldova''), and  
(iii) early termination or truncation.

While average WER and CER are low, we observe a standard deviation of approximately 7\%. This variance is largely attributable to the nature of the \textit{emg2qwerty} dataset, which includes short and syntactically sparse phrases. In such cases, even a single word error can disproportionately affect WER (e.g., 25\% WER for a 4-word sentence with one incorrect word). Nonetheless, the model generally preserves sentence length and fluency, even when lexical accuracy varies. Despite upstream errors introduced by finger misclassifications or variability in finger-to-key mappings, the transformer-based decoder consistently, hence supporting our modular framework.

\begin{tcolorbox}[colback=blue!5!white,colframe=blue!70!black,title=\textbf{Key Takeaway from RQ1}] On \textit{emg2qwerty}, MyoText reaches 85.4\% finger classification and 5.4\% sentence CER via a three-stage modular framework aligned with motor→ergonomic→linguistic structure: (1) \emph{Finger}: sEMG classifies fingers better than letters (direct letters: 51.2\%). (2) \emph{Ergonomics}: QWERTY finger–key constraints cut the per-position search depending on the standard finger-to-letter mapping. (3) \emph{Sentence Generation}: a T5 decoder resolves residual ambiguity, yielding coherent sentences without per-user calibration. \end{tcolorbox}



\subsection{RQ2: Generalization Under Real-World Typing and Sentence Variability}

A key requirement for practical sEMG-to-text decoding is the ability to generalize beyond the constraints of the training dataset. While \textit{emg2qwerty} serves as an important first benchmark for sEMG-to-text research, it remains limited in both linguistic diversity and typing variability. To assess broader generalization, we evaluate our approach in three settings: (i) sentence-level decoding on naturalistic, paragraph-scale text, (ii) robustness to deviations from canonical finger-to-key mappings, and (iii) cross-domain evaluation using open-source text datasets with simulated typing variability, since no open-source sEMG-to-sentence datasets currently exist.
\begin{wrapfigure}[13]{r}{0.45\textwidth}
\centering
\vspace{-6mm}
        \includegraphics[width=1\linewidth,height=5cm]{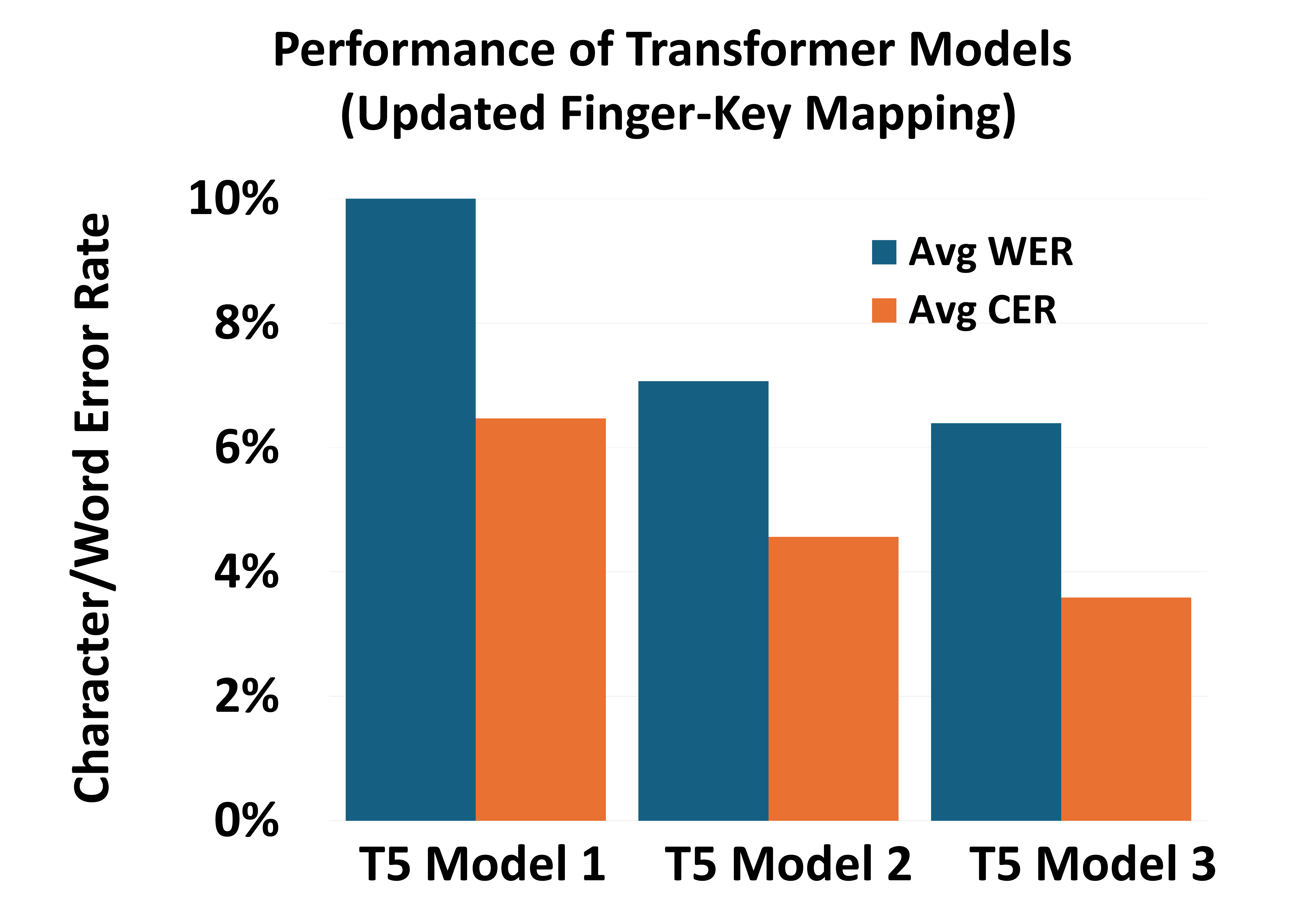}
        \vspace{-10mm}
        \caption{\textbf{Decoding Performance of T5 Variants as mentioned in RQ1  with Updated Letter Pool.} Lower the CER and WER indicates better performance. }
        \label{fig:t5_noise}
\end{wrapfigure}

\begin{figure}[t]
\centering
    \captionof{table}{\textbf{Comparisons between Reference and Predicted Sentences using Natural English Sentences.}}
    \includegraphics[width=0.8\linewidth]{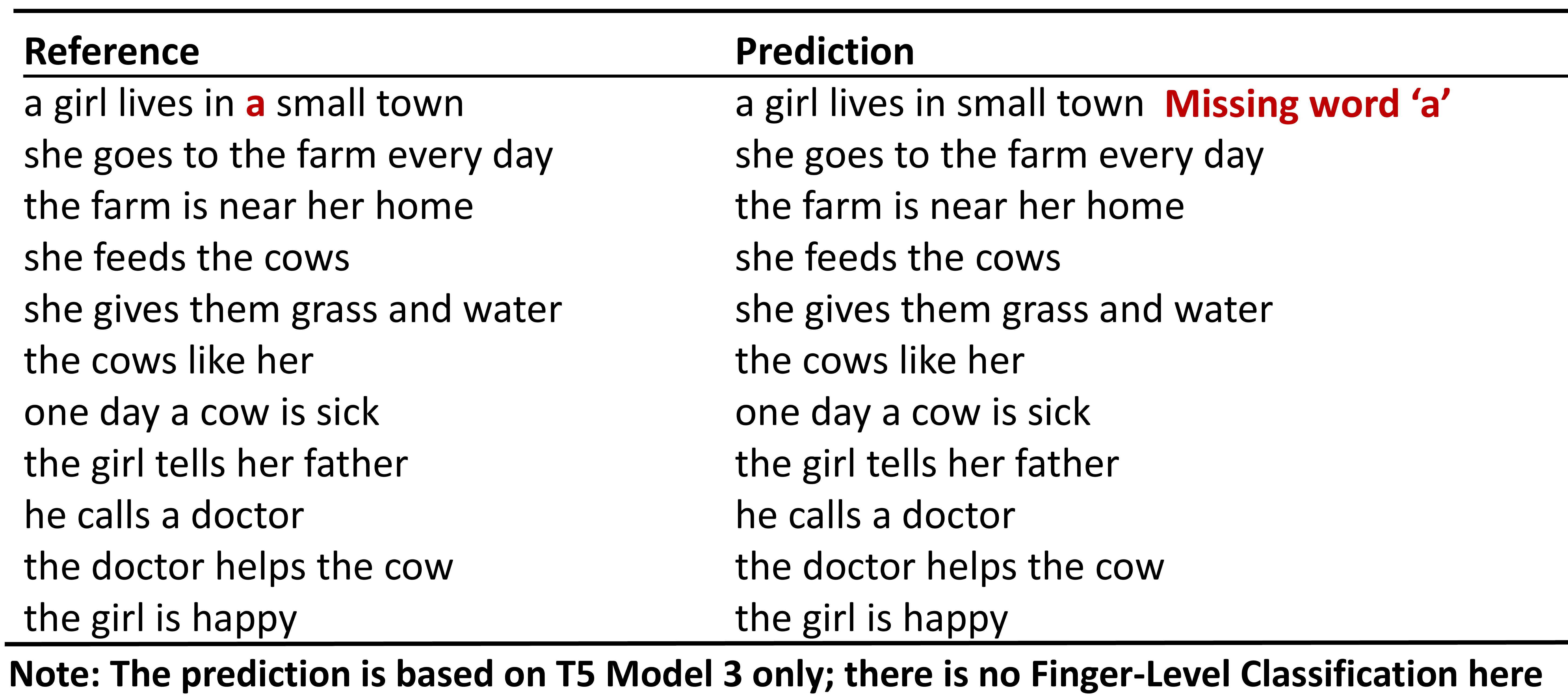}
    \label{tab:random_paragraph}
\end{figure}

\subsubsection{\textbf{Natural Language Generalization}}
To evaluate whether the sentence decoder overfits to the short, syntactically sparse structures in \textit{emg2qwerty} (mean sentence length $\approx$4 words), we tested the best-performing transformer model (T5 Model 3) on a paragraph of declarative English sentences, each of similar length, generated via ChatGPT-4. This evaluation isolates the language model's capability using candidate word pools only; no sEMG signals or finger classification are involved. As shown in Table~\ref{tab:random_paragraph}, the sentence decoder reconstructs long-form sentences with near-perfect fidelity. The only deviation is a minor article omission (``a girl lives in a small town'' $\rightarrow$ ``a girl lives in small town''), resulting in a WER of 1.43\% and CER of 0.41\%. These results confirm that the model generalizes well and does not memorize fixed phrase structures.

\subsubsection{\textbf{Robustness to Typing Variability}}
\label{sec:rq3_fingernoise}
While the QWERTY layout defines a canonical finger-to-key mapping, real-world typing often diverges, particularly for centrally located keys. For instance, the letter \texttt{b}, typically assigned to the left index finger (Finger 3), is frequently typed with the right index (Finger 6); similarly, \texttt{x}, \texttt{s}, and \texttt{w} may shift from the left ring (Finger 1) to the left middle finger (Finger 2), and \texttt{o}, \texttt{l}, and \texttt{p} are often typed with the right middle (Finger 7) instead of the right ring or pinky. To capture such variation, we augmented the candidate letter pools with plausible alternatives based on observed typing behavior (Table~\ref{tab:typing-variability-map}). This introduces realistic ambiguity into decoding, simulating behavioral noise. We re-evaluated all T5 models (Models 1–3) on \textit{emg2qwerty} using these expanded pools, keeping all other pipeline components unchanged. As shown in Fig~\ref{fig:t5_noise}, performance remains similar, with only minor increases in CER and WER. These findings confirm that \SYSTEM{} generalizes not only linguistically but also behaviorally, preserving decoding fidelity under realistic finger-to-key variation. This robustness supports its deployment across diverse user typing styles and input conditions.

\begin{table*}[t]
\caption{\textbf{Augmented Finger-to-Letter Mappings.} Canonical QWERTY pools are extended with additional letters to reflect plausible typing deviations, to simulate user variability, and evaluate robustness.}
  \small
  \label{tab:typing-variability-map}
  \begin{tabular}{p{3.2cm} p{4cm} | p{3.2cm} p{4cm}}
    \toprule
    \textbf{Finger (Left)} & \textbf{Updated Letter Pool} & \textbf{Finger (Right)} & \textbf{Updated Letter Pool} \\
    \midrule
    Pinky (0)  & \texttt[q, a, z] & Pinky (9) & \texttt[p] \\
    Ring (1)   & \texttt[w,s,x]       & Ring (8)  & \texttt[o, l] \\
    Middle (2) & \texttt[e,d,c,w,s,x] & Middle (7) & \texttt[i,k,o,l,p] \\
    Index (3)  & \texttt[r,f,v,t,g,b] & Index (6) & \texttt[y,h,n,u,j,m,t,g,b] \\
    \bottomrule
  \end{tabular}
\end{table*}

\subsubsection{\textbf{Cross-Dataset Generalization}}

\begin{wraptable}[7]{r}{0.4\textwidth}
\vspace{-2mm}
\centering
\caption{\textbf{Cross-Dataset Generalization.}}
\vspace{-2mm}
\label{tab:kaggle_cross}
\small
\begin{tabular}{c c c}
\toprule
\textbf{Train} & \textbf{Test} & \textbf{CER (\%)} \\
\midrule
D1 (741) & D2 (584) & 1.7 \\
D2 (584) & D1 (300) & 0.9 \\
\bottomrule
\end{tabular}
\end{wraptable}

A key application of our method is in AR/VR systems, such as Meta’s Neural Band and Ray-Ban Display~\cite{MetaRayB22:online}, where typing is primarily used for short-form communication (e.g., texting). To assess generalization beyond \textit{emg2qwerty}, we evaluated the sentence decoder on two public corpora from Kaggle~\cite{ChatSent12:online,Randomen9:online}, denoted D1 (chat sentiment; 741 sentences) and D2 (random English sentences; 584 sentences). While these datasets do not contain sEMG recordings, they are used to evaluate the sentence decoder’s linguistic generalization under simulated input variability. Finger-level perturbations, derived from the augmented mappings in Table~\ref{tab:typing-variability-map}, enable consistent benchmarking of decoding robustness. We applied the best-performing model (T5 Model~3) under these conditions and conducted cross-domain tests, where training was performed on D1 and evaluation on D2, then reversing the setup with a random subset of D1 reserved for testing (Table~\ref{tab:kaggle_cross}). This evaluation intentionally excludes \textit{emg2qwerty}, which was designed primarily to validate hardware-level interaction and thus contains mostly short (four-word) sequences with limited syntactic structure, which is appropriate for simple statistical baselines (e.g., $n$-grams). In contrast, D1 and D2 comprise longer, grammatically coherent sentences, providing a more challenging and representative testbed for cross-domain decoding. Across these conditions, the sentence decoder maintains low CER ($<\!2\%$) despite linguistic variability, indicating that the T5-based decoder is robust and well-suited for deployment beyond controlled lab settings.

\begin{tcolorbox}[colback=blue!5!white,colframe=blue!70!black,title=\textbf{Key Takeaway from RQ2}]
The \SYSTEM{} framework generalizes effectively beyond the \textit{emg2qwerty} dataset.  
It reconstructs longer, naturalistic sentences with high fidelity, remains robust under typing variability introduced by non-canonical finger-to-key mappings, and sustains performance when tested on unseen open-domain datasets.  
These results highlight the sentence decoder’s resilience to input noise and domain shift, underscoring its potential for deployment in diverse real-world scenarios.  
\end{tcolorbox}



\subsection{RQ3: Analyzing \textit{emg2qwerty} Dataset}

We conduct an in-depth analysis to validate the quality and relevance of the \textit{emg2qwerty} dataset used in our evaluation. It is a recently released large-scale dataset that has not yet been extensively studied. The dataset contains sEMG recordings collected with a wrist-mounted array while participants typed on a QWERTY keyboard. The corpus comprises sEMG and corresponding keystroke data from 108 participants, with prompts drawn from dictionaries and Wikipedia to ensure lexical and syntactic coverage. To assess suitability for our study, we analyze signal consistency and finger-specific patterns among users with high-quality tags from the dataset.

\begin{figure*}[b]
    \centering
    \captionsetup{font=small} 
    \begin{minipage}[t]{0.46\linewidth}
        \centering
        \includegraphics[width=\linewidth]{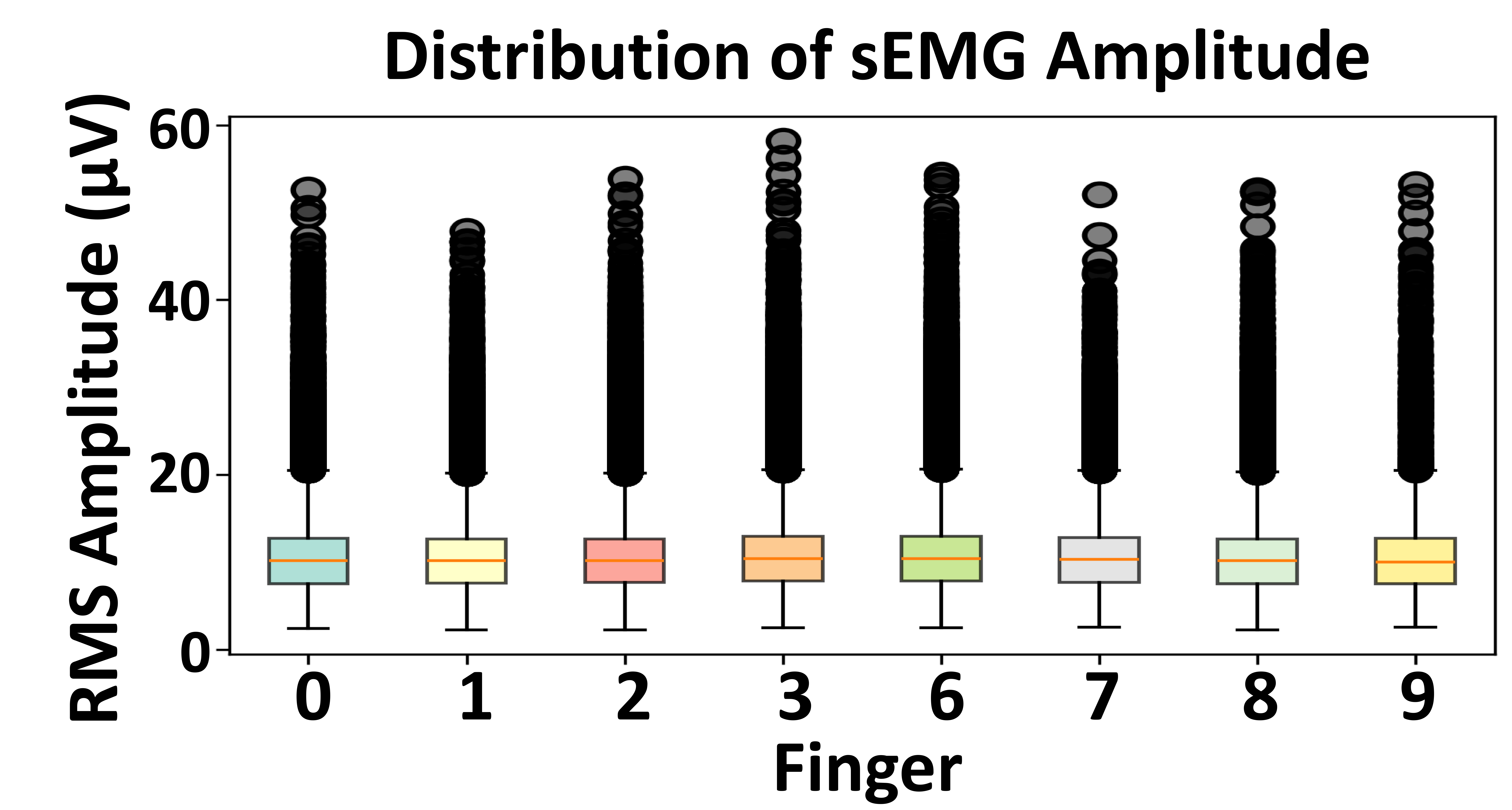}
    \caption{\textbf{sEMG amplitudes by finger class.} Computed from 2\,s windows centered on each keypress. Points above 40--50~$\mu$V correspond to key presses rather than outliers.}

        \label{fig:avg_rms_per_finger}
    \end{minipage}\hfill
    \begin{minipage}[t]{0.46\linewidth}
        \centering
        \includegraphics[width=\linewidth]{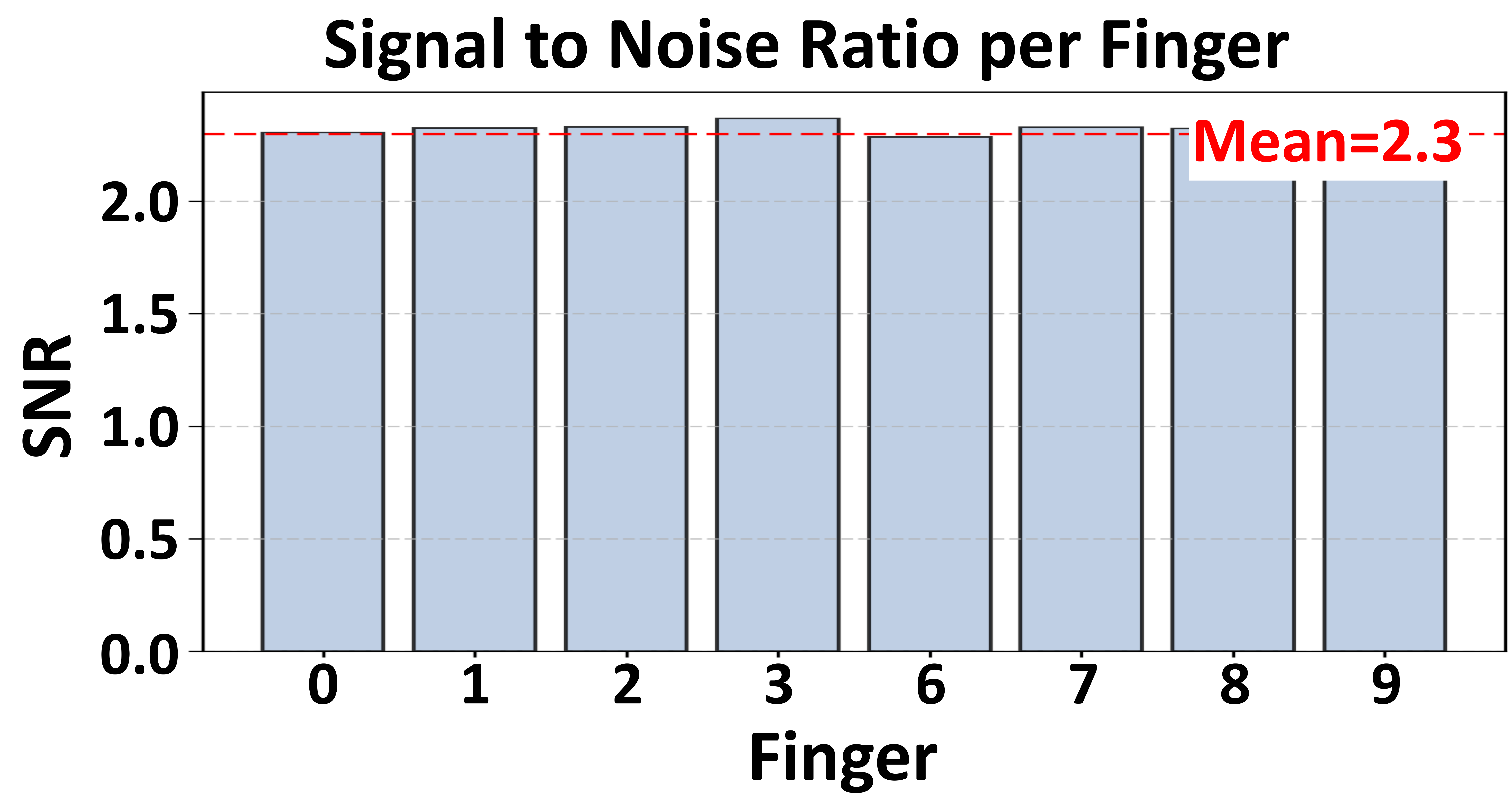}
        \caption{\textbf{Within-class RMS stability (}\(\mu/\sigma\)\textbf{).} Mean-to-std ratio per finger over 2\,s windows; values cluster near 2.3, indicating consistent signal quality.}
        \label{fig:snr_finger}
    \end{minipage}
    \vspace{-3mm}
\end{figure*}

\paragraph{\textbf{Signal quality and finger-level consistency}}
To verify signal stability, we computed RMS amplitude distributions for each finger over a 2\,s window centered on each keystroke (1\,s before and 1\,s after). This window captures both pre-activation and relaxation phases, which lowers the median amplitude relative to peak contraction values. As shown in Fig.~\ref{fig:avg_rms_per_finger}, median RMS amplitudes range from 10.6-11.0~$\mu$V with standard deviations of 4.6-5.1~$\mu$V. Interquartile ranges remain tight (approximately 5-15~$\mu$V), while higher-magnitude points (extending to 50--60~$\mu$V) correspond to key presses rather than statistical outliers. The coefficient of variation (CV = $\sigma_{\text{RMS}} / \mu_{\text{RMS}}$) ranges from 42-47\%, suggesting consistent scaling across fingers and systematic, not random, variability.

To quantify signal consistency, we calculated the signal-to-noise ratio (SNR) for each finger, defined as the inverse of the coefficient of variation ($\text{SNR} = \mu_{\text{RMS}} /\sigma_{\text{RMS}}$). As shown in Fig.~\ref{fig:snr_finger}, all fingers exhibit consistent and repeatable signals, with SNR values between 2.12 and 2.37 (mean=2.3). Finger 3 has the highest SNR (2.37), indicating highly repeatable muscle activations, likely due to its independent motor control~\cite{fuglevand1999force}. Finger 9 has the lowest SNR (2.12), which may result from its complex anatomy and great overlap with neighboring muscles~\cite{leijnse2010kinematic}.

These consistent amplitude distributions and SNR values confirm that the dataset captures stable, physiologically plausible muscle activation patterns with sufficient signal quality for reliable 
keystroke-level decoding

\begin{figure*}[t]
    \centering
    \includegraphics[width=\linewidth]{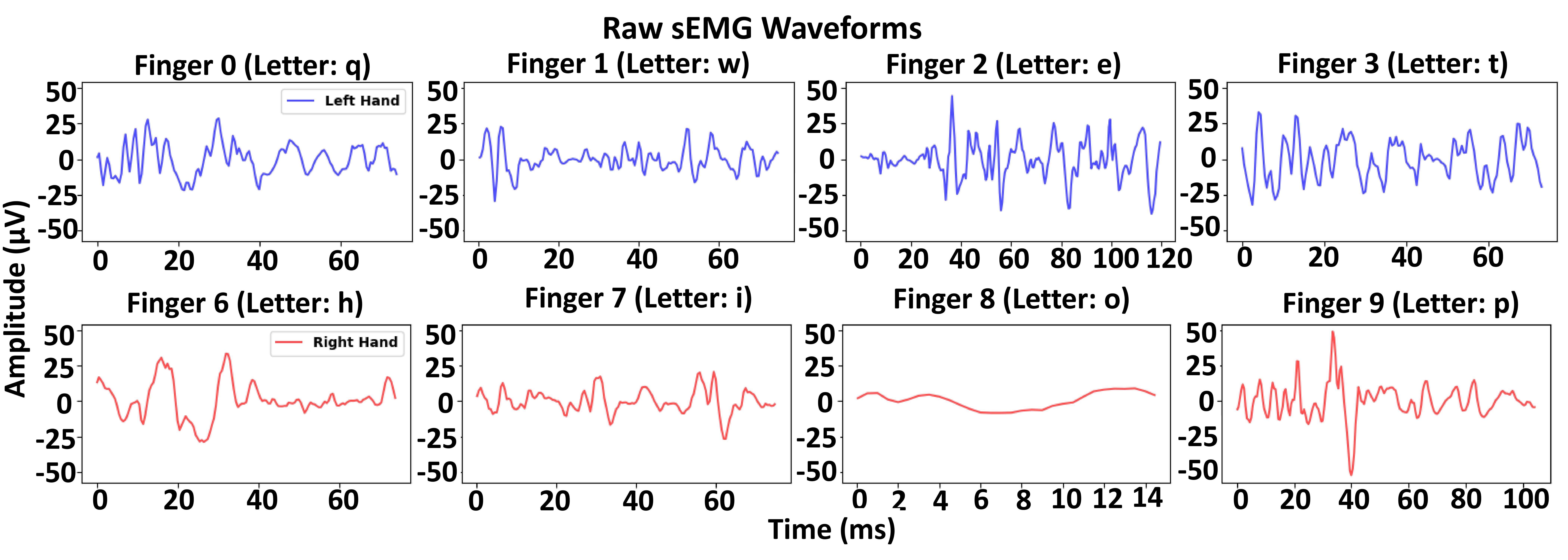}
    \vspace{-6mm}
    \caption{\textbf{Finger-specific sEMG waveforms.} Averaged single-press waveforms show distinct temporal morphologies across fingers, supporting temporally-aware decoding.}
    \label{fig:raw_emg_per_finger}
    \vspace{-2mm}
\end{figure*}

\paragraph{\textbf{Finger-specific waveform morphology.}}
Beyond aggregate amplitude metrics, we examine whether individual fingers produce distinguishable temporal patterns. Fig.~\ref{fig:raw_emg_per_finger} shows raw sEMG for single keypresses across fingers 0-9. Waveforms exhibit apparent finger-specific differences in duration (15-110 ms), peak amplitude ($\pm$20-50 $\mu$V), and frequency content. Finger 8 ('o') shows brief, smooth activation ($\sim$15 ms, $\pm$20 $\mu$V), while Finger 2 ('e') displays sustained high-frequency activity ($\sim$110 ms). Finger 9 ('p') reaches the highest amplitude ($\pm$50 $\mu$V), and Finger 6 ('h') shows dense transients. These distinct temporal signatures demonstrate finger-level discriminability and motivate temporal modeling approaches that capture waveform morphology rather than amplitude alone. These distinct temporal signatures qualitatively illustrate finger-level discriminability and motivate temporal modeling approaches that capture waveform morphology rather than amplitude alone.

\begin{tcolorbox}[colback=blue!5!white,colframe=blue!70!black,title=\textbf{Key takeaway for RQ3}]
The \textit{emg2qwerty} dataset exhibits consistent signal quality
and distinct finger-specific patterns in both amplitude and temporal domains. This stability and discriminability across users make it suitable for evaluating supervised keystroke decoding approaches.
\end{tcolorbox}



\subsection{RQ4: Positioning MyoText within sEMG-to-Text Systems and Real-World Deployment}

To evaluate the practical utility of \SYSTEM{}, we situate it within prior sEMG-to-text literature and examine how its modular design supports deployment on contemporary wearable platforms.

\paragraph{\textbf{Existing Approaches of sEMG-to-text Systems}}
Table~\ref{tab:emg_text_comparison} summarizes representative sEMG-to-text systems reported in the literature. While these studies differ in terms of datasets, vocabulary size, and task formulation, they collectively illustrate advancements in EMG-based text systems, ranging from gesture-driven interfaces to sentence-level decoding on compact wearable devices. Within this evolving landscape, \SYSTEM{} contributes an expanded evaluation scale, with both \SYSTEM{} and the \textit{emg2qwerty} baseline tested on $\geq$30 participants. This broader cohort supports stronger insights into generalizability and user adaptation compared to the smaller-scale evaluations common in earlier work.
On the \textit{emg2qwerty} dataset—using a wrist-worn sEMG device and the same 30-user configuration—\SYSTEM{} achieves a 5.4\% character error rate (CER), whereas the method in~\cite{sivakumar2024emg2qwerty} yields 64\% CER (i.e., 36\% per-character accuracy, since $\mathrm{CER}=1-\mathrm{accuracy}$) under identical conditions (Table~\ref{tab:emg_text_comparison}). This accuracy level aligns with our findings in RQ1 (\Cref{sec:rq1}), where letter classification achieved a similar accuracy range. Reported results in the broader literature span a range of system designs and evaluation setups. For example, \citet{wang2024efficient} report a 3.8\% CER on a 30-symbol set, and \citet{crouch2021natural} achieve 91\% accuracy using dual-forearm sleeves for podcast transcription. Together, these works delineate the current performance envelope of sEMG-based decoding systems under diverse vocabulary sizes, datasets, and device configurations. The reduction in CER observed for \SYSTEM{} reflects the efficacy of its modular framework, which integrates a physiological finger-motion classifier with a transformer-based sentence decoder. By decoupling physiological decoding from linguistic reasoning, the approach generalizes across users.

\begin{table}[t]
  \centering
\caption{\textbf{sEMG-to-Text Landscape.} Representative systems reported in the literature. Metrics are shown as reported: Accuracy (Acc) indicates the fraction of correctly predicted characters, while Character Error Rate (CER) represents the fraction of incorrectly predicted characters (\( \text{CER} = 1 - \text{Acc} \)).}
  \small
  \label{tab:emg_text_comparison}
  \begin{tabularx}{\linewidth}{l l l l Y l}
    \toprule
    \textbf{System} & \textbf{Users} & \textbf{Channels} & \textbf{Device} & \textbf{Task} & \textbf{Metric} \\
    \midrule
    Sharma et al.~\cite{sharma2017neural} & 10 & 2 & BIOPAC & 7 keys  & 92.7\% Acc \\
    Fu et al.~\cite{fu2020typing} & -- & 8 & Myo band & T9 gestures (9 keys) & 85.9\% Acc \\
    Crouch et al.~\cite{crouch2021natural}$^{\ddagger}$ & 148k keys$^{\ddagger}$ & 32 & Dual sleeves & Podcast transcription (full vocab) & 91\% Acc \\
    Wang et al.~\cite{wang2024efficient} & -- & 32 & 16$\times$16 band & 30 symbols (26 letters + space/backspace/enter/period) & 3.8\% CER \\
    Sivakumar et al.~\cite{sivakumar2024emg2qwerty}$^{\dagger}$ & 108 & 32 & Wrist band & Full sentences (26 keys,$\sim$4 words per sentence) & 52\% CER  \\
    Sivakumar et al.~\cite{sivakumar2024emg2qwerty}$^{\dagger}$ & 30$^{*}$ & 32 & Wrist Band & Full-sentence decoding (26 keys, $\sim$4 words/sentence) & 64\% CER (36\% Acc)\\
    \textbf{\SYSTEM{} (ours)$^{\dagger}$} & \textbf{30} & \textbf{32} & \textbf{Wrist band} & \textbf{Full sentences (26 keys, $\sim$4 words per sentence)} & \textbf{5.4\% CER (94.6\% Acc)} \\
    \bottomrule
  \end{tabularx}
  \vspace{2mm}
\begin{minipage}{\linewidth}
\footnotesize
\raggedright
\textit{Note:}
User cohorts are shown as reported (— = not specified). 
$^{\dagger}$ denotes evaluation on the same dataset and hardware (\textit{emg2qwerty}, 32-channel wrist band). 
$^{*}$ indicates reimplementation and re-evaluation on the 30-user subset used for \SYSTEM{}. 
$^{\ddagger}$ indicates keystrokes reported instead of participant count. 
\par
\end{minipage}
  \vspace{-8mm}
\end{table}

\paragraph{\textbf{Deployment Feasibility}}
Beyond accuracy, the practical viability of sEMG-to-text systems depends on computational efficiency and real-time operation within the strict power and memory constraints of wearable devices. A monolithic end-to-end model that jointly decodes finger movements and generates text would exceed the compute and power budgets typical of wearables. Conversely, a cloud-only approach introduces privacy risks, network dependence, and latency~\cite{kisaco2024edge,fiveable2024power}. To address these challenges, \SYSTEM{} adopts a split-execution design where the finger classifier can operate locally on the wearable device, while the transformer-based sentence decoder can run on a paired phone or edge processor.
Profiling the two main components of \SYSTEM{} shows that Step 2 (finger classifier) executes in $\sim$60ms per sample, requiring only $\approx$12MOPs and $<12$MB of memory, which is well within interactive thresholds for contemporary wearables~\cite{helios2024gesture,sayeed2023edge,wang2024efficient}. Modern edge NPUs deliver 38–80TOPS (Apple M4, Snapdragon X2 Elite)~\cite{apple2024m4specs,techradar2024m4tops,windowscentral2025snapdragon,tomsguide2025snapdragon}, readily supporting this module on-device. Step 4 (transformer decoder) is computationally more intensive but is efficiently handled by mobile-class hardware. On an NVIDIA L4 GPU, inference runs at 18.851ms/token. Advances in quantization, kernel fusion, and attention caching now enable billion-parameter transformers to operate on mobile NPUs~\cite{arxiv2025kernels,meta2024quantized,plainenglish2025mobile}. Commercial deployments (e.g., Apple ANE, Google on-device LLMs) already demonstrate low-latency text generation without that in cloud-independent.

Together, these results indicate that \SYSTEM{} is deployable on current hardware: the finger classifier runs locally on the resource-constrained wearable, while the language model executes on a paired, resource-rich device such as a smartphone or edge processor.

\begin{tcolorbox}[colback=blue!5!white,colframe=blue!70!black,title=\textbf{Key Takeaway from RQ4}]
\SYSTEM{} achieves an accuracy (5.4\% CER) while offering key advantages, such as cross-user generalization and a physiologically grounded architecture. The finger classifier fits within on-wrist compute budgets, and the transformer decoder runs efficiently on paired smartphones, enabling practical, privacy-preserving sEMG text entry on current hardware.
\end{tcolorbox}

\section{Discussion and Future Work} \label{sec:discussion}

\SYSTEM{} introduces the first physiologically grounded approach to sEMG-based text decoding by explicitly modeling the relationship between muscular activation and typing behavior. Unlike prior end-to-end systems that directly map sEMG signals to letters, \SYSTEM{} first predicts the active finger, which is a feature directly linked to EMG activity, before invoking a transformer-based language model for sentence decoding. This hierarchical design substantially reduces the effective search space of possible letters enabling the language model to operate within well-defined, context-aware constraints. The result is a decoding framework that couples physiological interpretability with linguistic fluency, achieving low character error rates while preserving computational efficiency suitable for wearable deployment.

However, several opportunities exist to strengthen our findings. First, while our evaluation uses a single dataset, it represents the most extensive available sEMG typing resource with 108 participants. Due to computational constraints, we evaluated our model on 30 users, which is a robust sample size that is larger than those used in many previous EMG typing studies. This sample size is justified by our modular framework, and the LOUO validation confirms consistent cross-user generalization. Our 30-user cohort with explicit LOUO evaluation matches or surpasses prior sEMG-to-text work (see Table~\ref{tab:emg_text_comparison}). In future work, we will evaluate the full participant pool to validate generalizability further. Second, this work focuses on validating the core innovation of the \SYSTEM{} framework rather than its end-to-end deployment. Future implementations incorporating real-time sEMG acquisition and on-device processing will provide crucial insights into system latency and power consumption for practical wearable deployment. Third, while we employ standardized QWERTY finger-key mappings, individual typing strategies vary considerably. Our experiments incorporate noise models to simulate these deviations, and the \textit{emg2qwerty} dataset's trained typists exhibit more consistent patterns than general users. A comprehensive survey of population-level finger-key preferences would enable more personalized decoding strategies. \SYSTEM{} establishes a viable foundation for hands-free text input; the challenge ahead is scaling the data, refining the models, and validating the system outside controlled settings. \SYSTEM{}'s flexibility suggests broad applications, including gesture interfaces, assistive technologies for motor impairments, and silent communication in noise-sensitive environments, providing a blueprint for next-generation biosignal-based human-computer interaction.

\section{Conclusion}\label{sec:conclusion}

We presented \SYSTEM{}, a physiologically grounded and modular sEMG-to-text decoding framework that integrates finger-level classification, ergonomically constrained letter mapping, and a transformer-based sentence generation model. Unlike prior monolithic or gesture-based systems, \SYSTEM{} establishes a structured motor–linguistic hierarchy that mirrors the neuromuscular process of typing. Leveraging the \textit{emg2qwerty} dataset, \SYSTEM{} achieves 85.4\% finger-classification accuracy and 5.4\% character error rate (CER), demonstrating reliable decoding performance across users. This work represents a step toward scalable, keyboard-free communication interfaces, advancing the practical realization of neural input for AR/VR environments, assistive technologies, and ubiquitous computing platforms.



\bibliographystyle{ACM-Reference-Format}
\bibliography{ref}

\end{document}